\documentclass[conference]{IEEEtran}
\usepackage{tikz}
\usepackage{amsmath}

\usepackage{comment}
\usepackage{cite}
\usepackage{amsmath,amssymb,amsfonts}
\usepackage{algorithmic}
\usepackage{textcomp}
\usepackage{float}
\usepackage{xspace}
\usepackage{makecell}
\usepackage{fixltx2e}
\usepackage{fancyhdr}
\usepackage{lipsum}
\usepackage{soul,xcolor}
\usepackage{xparse}
\usepackage[colorlinks,
        citecolor=blue,
        linkcolor=blue,
        urlcolor=black]{hyperref}

\usepackage{filecontents}
\newif\ifdraft
\draftfalse

\newif\ifpreprint
\preprinttrue
\setstcolor{blue}
\ifdraft
  \newcommand{\todocolor}[1]{\textcolor{red}{#1}}
  \newcommand{\todocolors}[1]{\textcolor{teal}{#1}}
  \newcommand{\todocolorc}[1]{\textcolor{blue}{#1}}
  
\else
  \newcommand{\todocolor}[1]{}
  \newcommand{\todocolors}[1]{}
  \newcommand{\todocolorc}[1]{}
  
\fi
\newcommand{\mahmood}[1]{\todocolor{[[Mahmood: #1]]}}
\newcommand{\weiran}[1]{\todocolors{[[Weiran: #1]]}}

\newcommand{\keane}[1]{\todocolor{[[Keane: #1]]}}

\newcommand{\figref}[1]{Fig.~\ref{#1}}
\newcommand{\figrefs}[2]{Figs.~\ref{#1}--\ref{#2}}
\newcommand{\tabref}[1]{Fig.~\ref{#1}}
\newcommand{\secref}[1]{\S\ref{#1}}

\newcommand{\eqnref}[1]{Eqn.~\ref{#1}}

\newcommand{\mdtrain}{\text{MDTRAIN}\xspace}
\newcommand{\mdgroup}{\text{MDMUL}\xspace}
\newcommand{\mdmax}{\text{MDMAX}\xspace}
\newcommand{\mdgrouploss}{the \mdgroup{} loss\xspace}
\newcommand{\mdmaxloss}{the \mdmax{} loss\xspace}
\newcommand{\elmdmax}{\ensuremath{\ell_{\mathit{\mdmax}}}}
\newcommand{\elmdgroup}{\ensuremath{\ell_{\mathit{\mdgroup}}}}
\newcommand{\md}{MD\xspace}
\newcommand{\imagenet}{ImageNet\xspace}
\newcommand{\sst}{SST-5\xspace}
\newcommand{\tpgd}{T-PGD\xspace}
\newcommand{\pubfig}{PubFig\xspace}
\newcommand{\gtsrb}{GTSRB\xspace}
\newcommand{\autopgd}{Auto-PGD\xspace}
\newcommand{\doa}{DOA\xspace}
\renewcommand{\paragraph}[1]{\subsubsection{#1}}

\newcommand{\prob}[1]{\ensuremath{\mathbb{P}\left({#1}\right)}\xspace}

\newcommand{\groundTruth}{\ensuremath{f}\xspace}
\newcommand{\classifier}{\ensuremath{\tilde{f}}\xspace}
\newcommand{\instance}{\ensuremath{x}\xspace}
\newcommand{\instanceAlt}{\ensuremath{x'}\xspace}
\newcommand{\instanceUniverse}{\ensuremath{\mathcal{X}}\xspace}
\newcommand{\instanceSetGen}{\ensuremath{\mathcal{G}}\xspace}
\newcommand{\instanceSet}{\ensuremath{X}\xspace}
\newcommand{\classUniverse}{\ensuremath{\mathcal{Y}}\xspace}
\newcommand{\perturbations}{\ensuremath{R}\xspace}
\newcommand{\allowedPert}{\ensuremath{\Pi}\xspace}
\newcommand{\impersonationSet}{\ensuremath{\tilde{I}}\xspace}
\newcommand{\impersonationGoal}[1]{\ensuremath{I_{#1}}\xspace}
\newcommand{\impersonationGoalSet}{\ensuremath{\mathcal{I}}\xspace}
\newcommand{\impersonationGoalIdx}{\ensuremath{i}\xspace}
\newcommand{\impersonationGoalNmbr}{\ensuremath{m}\xspace}
\newcommand{\adversary}{\ensuremath{A}\xspace}
\newcommand{\sourceclass}{\ensuremath{s}\xspace}
\NewDocumentCommand{\targetclass}{ g }{\ensuremath{t\IfNoValueF{#1}{_{#1}}}\xspace}
\newcommand{\targetclassset}{\ensuremath{T}\xspace}
\newcommand{\target}{\ensuremath{t}\xspace}
\newcommand{\sourceclassset}{\ensuremath{S}\xspace}

\newcommand{\advtrainweight}{\ensuremath{\kappa}\xspace}

\newcommand{\impRel}{\textsf{imp-rel}\xspace}
\newcommand{\experiment}[2]{\ensuremath{\mathbf{Expt}^{#1}_{#2}}\xspace}
\newcommand{\advantage}[2]{\ensuremath{\mathbf{Adv}_{#2}}\xspace}
\newcommand{\robustness}[1]{\ensuremath{\mathbf{Rob}_{#1}}\xspace}

\newcommand{\gbtrfull}{group-based robustness\xspace}
\newcommand{\Gbtrfull}{Group-based robustness\xspace}
\newcommand{\gbtrfullCAPS}{Group-based Robustness\xspace}
\newcommand{\gbtrapprev}{GBR\xspace}

\newcommand{\gbtafull}{group-based attacks\xspace}

\newcommand{\dnn}{DNN\xspace}

\ifCLASSINFOpdf
\else
\fi
\usepackage{tcolorbox}
\tcbuselibrary{skins,breakable}
\usetikzlibrary{shadings,shadows}

\newenvironment{myblock}[1]{%
    \tcolorbox[beamer,%
    noparskip,breakable,
    colback=gray!40,colframe=gray,%
    colbacklower=gray!40,%
    title=#1]}%
    {\endtcolorbox}

\makeatletter
\newcommand{\linebreakand}{%
  \end{@IEEEauthorhalign}
  \hfill\mbox{}\par
  \mbox{}\hfill\begin{@IEEEauthorhalign}
}
\makeatother

\begin{document}
%
\title{Group-based Robustness:\\
 A General Framework
 for Customized Robustness\\ in the Real World}

\author{
\IEEEauthorblockN{Weiran Lin}
\IEEEauthorblockA{Carnegie Mellon University\\
weiranl@andrew.cmu.edu}
\and
\IEEEauthorblockN{Keane Lucas}
\IEEEauthorblockA{Carnegie Mellon University\\
kjlucas@andrew.cmu.edu}
\and
\IEEEauthorblockN{Neo Eyal}
\IEEEauthorblockA{Tel Aviv University\\
neoedan@gmail.com}
\linebreakand
\IEEEauthorblockN{Lujo Bauer}
\IEEEauthorblockA{Carnegie Mellon University\\
lbauer@cmu.edu}
\and
\IEEEauthorblockN{Michael K. Reiter}
\IEEEauthorblockA{Duke University\\
michael.reiter@duke.edu}
\and
\IEEEauthorblockN{Mahmood Sharif}
\IEEEauthorblockA{Tel Aviv University\\
mahmoods@tauex.tau.ac.il}}

\IEEEoverridecommandlockouts
\makeatletter\def\@IEEEpubidpullup{6.5\baselineskip}\makeatother
\IEEEpubid{\parbox{\columnwidth}{
    Network and Distributed System Security (NDSS) Symposium 2024\\
    26 February - 1 March 2024, San Diego, CA, USA\\
    ISBN 1-891562-93-2\\
    https://dx.doi.org/10.14722/ndss.2024.24084\\
    www.ndss-symposium.org
}
\hspace{\columnsep}\makebox[\columnwidth]{}}

\maketitle

\begin{abstract}
Machine-learning models are known to be vulnerable to evasion
attacks, which perturb model inputs to induce 
misclassifications. In this work, we identify real-world
scenarios where the threat cannot be assessed accurately by
existing attacks.
Specifically, we find that conventional metrics measuring targeted and
untargeted robustness do not appropriately reflect a model's ability to
withstand attacks from one \textit{set of source classes} to another \textit{set of
  target classes}.
To address the shortcomings of
  existing methods,
we formally define a new metric, termed
\emph{\gbtrfull}, that complements existing metrics and is
better-suited for evaluating model performance in certain attack
scenarios. We show empirically that \gbtrfull allows us to distinguish
between machine-learning models' vulnerability against specific threat
models in situations 
where traditional robustness metrics do not apply.
Moreover, to measure \gbtrfull efficiently and accurately,
we 1) propose two loss functions 
and 2) identify three new attack strategies. 
We show empirically that, with comparable success rates, finding
evasive samples using our new loss functions 
saves computation by a factor as large as the number of targeted classes,
and that finding evasive samples, using our new
attack strategies, saves time by up to 99\%
compared to brute-force search methods. 
 Finally, we propose a
defense method that  
increases \gbtrfull by up to 3.52 times.
\end{abstract}


%
\section{Introduction}
\label{sec:intro}
Machine-learning models are known to be vulnerable to evasion attacks---attacks that, by
slightly perturbing the models' input, cause models to
misclassify~\cite{ECML13:evasion}.  
Research that evaluates the susceptibility of models to evasion
attacks typically measures these models' classification accuracies
with benign and evasive inputs; these metrics are commonly referred to as benign accuracy and untargeted robustness, respectively~\cite{arxiv21:robustbench,iclr20:DSLH20,iclr20:WRK20,icml19:HLM19,iclr20:WZYBMG20,NeurIPS20:SWMJ20,NeurIPS19:CRSLD19,NeurIPS20:WXW20,NeurIPS20:SIEKM20,oakland19:DPDefense,icml19:CSmooth,icml19:Provable,popl19:DeepPoly,arxiv17:Detect,arxiv20:CompareDefenses,iclr18:JPEG,iclr18:DGAN,NeurIPS21:GRWSCM21,NeurIPS21:KSDT21,ICML22:PLYZY22,ICLR22:SMHDXCM22,NeurIPS21:HWEGBM21,ACC22:SSLW22,ICLR21:ZZNHSK21,NeurIPS20:AF20,NeurIPS19:ZZLZD19,CVPR20:CLCCAW20,ICCV21:CLWJ21,ICML19:ZYJXGJ19,NeurIPS20:HZZ20,ICML20:RWK20,ICML20:ZXHNCSK20,ECCV20:AMH20}.
Previous work also defines models' \emph{targeted robustness} as their ability
to resist making specific (mis)classifications when faced with
evasion attempts~\cite{GlobalSIP18:targeted,ATVA18:targeted,ICML22CGD}.

\begin{figure}[t!]
\centerline{\includegraphics[width=0.5\columnwidth]{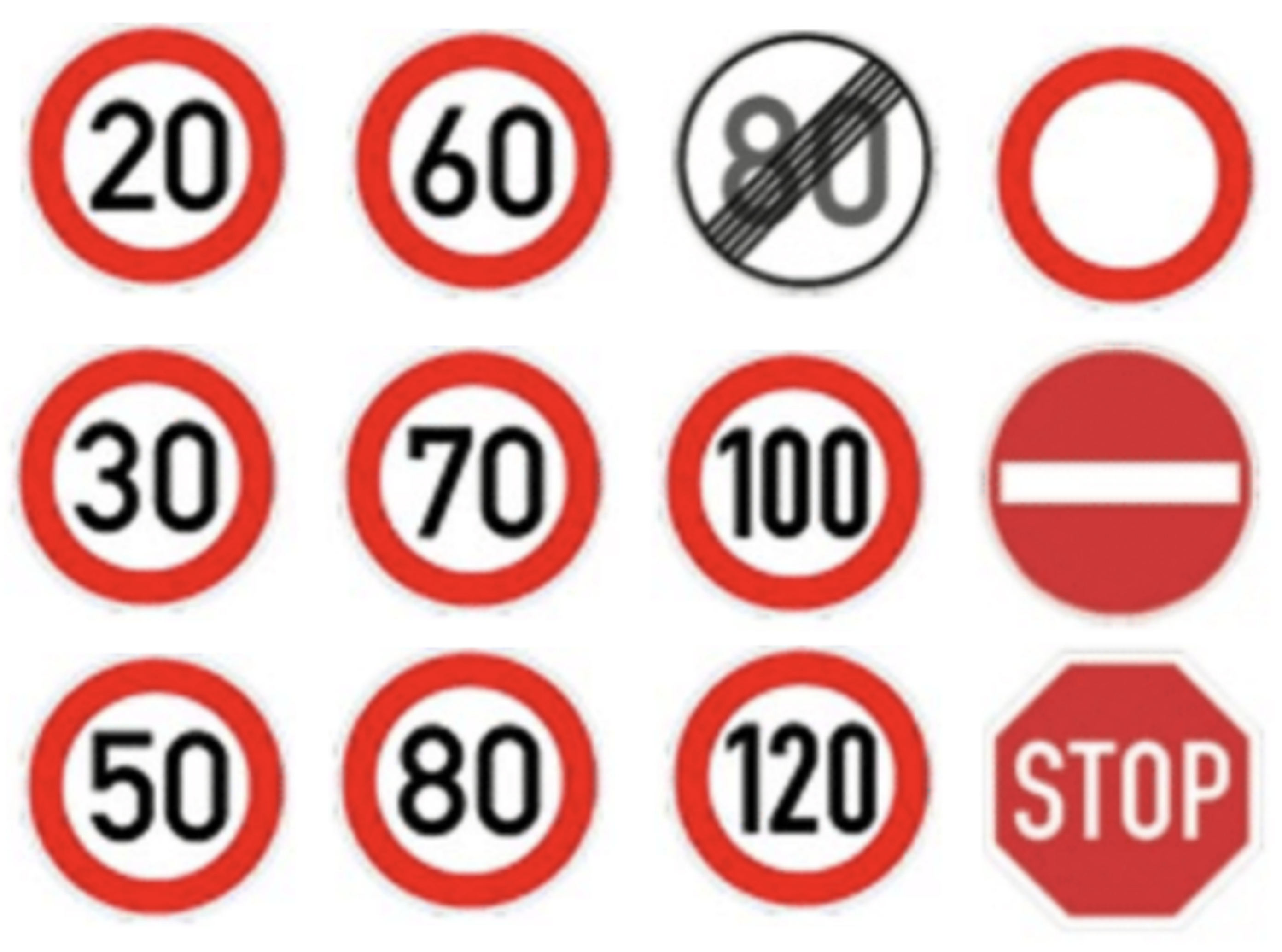}}
\caption{Traffic signs from GTSRB~\cite{IJCNN13:GTSRB}. The left three
  columns are speed-limit and delimit signs (i.e., ones that restrict speed limit or mark the end of restrictions). The rightmost column
  includes three signs that signify an immediate stop: no vehicles,
  no entry, and stop (from top to bottom).
  }
\label{fig:signs}
\end{figure}

However, more complicated threats exist in the real world. For
example, suppose adversaries want to induce traffic congestion or
self-driving vehicle accidents. 
Such adversaries could attempt to achieve
their goal by suddenly reducing the speed of certain vehicles, so that these vehicles might be hit by the vehicles behind them.
To do so, they might perturb a
\emph{specific} group of traffic signs,
such as speed limit and delimit signs, which restrict allowed speeds or remove
such restrictions (shown in \figref{fig:signs})\mahmood{not sure how fig 1 supports the example}\weiran{moved here}).
They might perturb these signs to signs that command an immediate stop, such as stop signs, no entry signs, and no vehicle
signs; or they could also perturb these signs into signs that display a limit much lower than the actual limit
(e.g., no more than half of the actual limit.
Adversaries achieve
their goal if they perturb the speed limit and delimit signs 
to be incorrectly classified as 
\emph{any} sign that requires an immediate stop or specifies a speed limit much lower than the actual limit. 

As another example, suppose a different adversary—a group of burglars, for instance—wants to illegally open a vault at a bank.
The bank requires three \emph{distinct} staff members to
give their permission before the vault can be opened; none of the members can open the vault alone. 
The group of burglars might therefore be able to 
succeed in opening the vault if they are able to impersonate any three distinct
individuals who work for the bank.
However,
they might be restricted to only a few attempts before triggering an alarm,
and thus they need attempt-efficient strategies for choosing the staff members that they would impersonate.

 We observe
that existing metrics---benign accuracy, untargeted robustness,
and targeted robustness---do not accurately measure models' ability
to resist making misclassifications in these examples and other similar
scenarios. 
Specifically, in the traffic sign example, benign accuracy and
untargeted robustness measure models' ability to resist predicting
\emph{any} inputs as \emph{any} incorrect classes. Targeted robustness
measures models' ability to resist predicting \emph{any} inputs as a
\emph{specific} incorrect class. None of the three metrics assess models'
ability to resist predicting inputs from one \emph{set of classes} as another,
mutually exclusive \emph{set of classes}.    
Additionally, in the burglary example, targeted and untargeted robustness
evaluate models on
a per-input-instance 
basis, 
while the models' ability to resist giving authorized access cannot be measured on a per-input-instance basis:
to open the vault, \emph{multiple ($>$1)} burglars may impersonate authorized bank staff simultaneously.
Hence, there is a need for
a new metric to better evaluate the susceptibility of models to
such threats. To this end, we formally
define this new metric, \emph{\gbtrfull}, as a model's ability to
resist attempts to cause
\emph{specific misclassifications} on data points \emph{from certain
classes}.
We then empirically demonstrate that this metric gives us insight into
models that previous metrics do not:
models that appear similar according to existing
metrics are actually very different by this new metric, and hence not
equally suitable in scenarios where robustness is essential
(\secref{sec:metric}).

While existing attacks can be used to estimate group-based robustness, they are inefficient at doing so.
As another contribution, we designed
more computationally efficient attacks, termed \emph{\gbtafull}, to help
compute \gbtrfull faster while attaining a comparable or higher level of
accuracy than the following na\"{i}ve methods:
\begin{itemize}
\item One possible na\"{i}ve method to perform \gbtafull is to attempt each of
the specified misclassifications on each input instance.
In the traffic sign example, 
attackers might launch three
individual targeted attacks to perturb a 60~KPH speed limit sign.
Each attack would try
 to perturb the sign
into one of the three signs that require an immediate stop, a 20~KPH speed limit, or a 30~KPH speed limit.
This approach
tends to find the most adversarial examples, and we use it to measure
\gbtrfull. However, compared with running one targeted
attack, this approach runs a set of targeted attacks and is more time-consuming.

\item Another possible na\"{i}ve
approach is to randomly select a single target class 
from a
specified set and perform standard targeted attacks.
For example, attackers may launch a single targeted attack in which
the target class is randomly selected from among the signs that require an immediate stop or display a limit much lower than the actual limit. 
While this
approach costs less in time than the previous one,
it tends to be significantly less successful.
\end{itemize}

To more quickly find perturbations that can cause misclassifications
among a specified set of candidates, we define two new loss functions,
$\ell_{\mdmax}$ and $\ell_{\mdgroup}$.
We empirically
verify that the loss functions boost the efficiency of attacks in scenarios like the traffic sign example (\secref{sec:attack:loss:results}).
Compared with iterating over all target classes to perform targeted attacks,
attacks with our loss functions were computationally cheaper
 by a factor as large as the number of targeted classes
 while still finding similarly many 
 successful perturbations.
Compared with randomly selecting a target class for each input instance to perform targeted
attacks, 
attacks with our loss functions were equally fast but found
successful perturbations up to 15$\times$ more often (\secref{sec:attacks:loss}).

Next, we propose more efficient attack strategies for settings
in which an attacker has a small set of inputs at their disposal and is
attempting to target a specific subset of classes.
In particular,
we define three
new attack strategies that
choose which misclassification to attempt by first estimating the individual chances of
success of perturbing each input instance into each target class.
In the burglary scenario, this would allow burglars to make fewer
impersonation attempts (or to better choose which subset of burglars
attempts the break-in)---the strategies estimate each burglar's chance
to impersonate each staff member and then attempt to cause
impersonations only for the most promising pairs. 
 We demonstrate that our new attack strategies 
boost the efficiency of attacks; e.g., in the burglary scenario,
compared with randomly selecting burglars and staff to launch attacks,
burglars would need up to 99\% fewer attack attempts with
these strategies (\secref{sec:attacks:strategy}).

Finally, we show how formalizing the real threat allows more effective
defenses against it: we demonstrate how to modify adversarial training
to increase \gbtrfull,
without losing benign accuracy or accuracy on classes that might be
impersonated (\secref{sec:defenses}). 
For example, in the burglary scenario, a face-recognition system
with our defense obtains up to 3.52$\times$ better robustness, with
  similar benign accuracy for all identities and
  similar benign accuracy for all staff members,
compared to existing defenses. 
We modified adversarial training to optimize these three metrics instead of conventional robustness metrics (\secref{sec:defenses}).


In summary, our contributions are the following:
\begin{itemize}
\item We define a new metric that better reflects many practical attack
  scenarios and more accurately evaluates their corresponding threat (\secref{sec:metric}).
\item We propose two loss functions that help attacks find
  misclassifications within a given set of targeted
  classes markedly faster than existing methods (\secref{sec:attacks:loss}).
\item We develop three attack strategies that when used individually
  or together, can produce diverse misclassifications for a given
  number of input instances with better time efficiency
  than brute-force approaches (\secref{sec:attacks:strategy}).
  \item We implement a defense method that improves the robustness of machine learning models
  against the attacks mentioned above (\secref{sec:defenses}).
\end{itemize}

Next, in \secref{sec:metric}, we further
motivate and formally define the new group-based metric. We introduce new
loss functions and new attack strategies that take advantage
of our metric in \secref{sec:attacks}. 
We also propose a defense that boosts the performance of models in this metric
in \secref{sec:defenses}.
Finally, we position this work in an overview of related
work in \secref{sec:relatedwork} and conclude in
\secref{conclusion}.

\section{\gbtrfullCAPS: A New Metric}
\label{sec:metric}
In this section, we introduce \gbtrfull, a new metric to
evaluate machine-learning models.
We first introduce existing evasion attacks and how robustness is typically measured (\secref{sec:metric:background}).
Then, we 
present real-world scenarios that demonstrate the importance of
this new metric (\secref{sec:metric:motivation}). Next, we
formally define the new metric, \emph{\gbtrfull}, and corresponding attacks, \emph{\gbtafull}
(\secref{sec:metric:definition}); and 
we show that \gbtafull constitute a broader space of evasion
attacks than had previously been studied (\secref{sec:metric:space}). 
Finally, we discuss
our experiments (\secref{sec:metric:setup}) and empirically demonstrate that \gbtrfull offers a meaningful
  assessment of model susceptibility to attacks in the real world
  that is orthogonal to conventional metrics 
(\secref{sec:metric:results}).

\subsection{Background}
\label{sec:metric:background}
Evasion attacks perturb the input of machine-learning models to induce
misclassifications. There are many implementations of evasion
attacks~\cite{icml20:autopgd,iclr18:PGD,iclr15:fgsm,iclr17:i-fgsm,iclr14:lbfgs,cvpr16:deepfool,cvpr18:momentum,Euro16:JSMA,ICMLA17:semantic,nips19:sparse,ICML18:spsa,ccs16:eyeglasses,Oakland17:CarliniWagner,iclr16:featureadv,Oakland20:HopSkip,ICML18:noise,ICML22CGD,ICAIIC19:multitarget,IEEE18:multitarget}, along with defenses against these
attacks~\cite{arxiv21:robustbench,iclr20:DSLH20,iclr20:WRK20,icml19:HLM19,iclr20:WZYBMG20,NeurIPS20:SWMJ20,NeurIPS19:CRSLD19,NeurIPS20:WXW20,NeurIPS20:SIEKM20,oakland19:DPDefense,icml19:CSmooth,icml19:Provable,popl19:DeepPoly,arxiv17:Detect,arxiv20:CompareDefenses,iclr18:JPEG,iclr18:DGAN,NeurIPS21:GRWSCM21,NeurIPS21:KSDT21,ICML22:PLYZY22,ICLR22:SMHDXCM22,NeurIPS21:HWEGBM21,ACC22:SSLW22,ICLR21:ZZNHSK21,NeurIPS20:AF20,NeurIPS19:ZZLZD19,CVPR20:CLCCAW20,ICCV21:CLWJ21,ICML19:ZYJXGJ19,NeurIPS20:HZZ20,ICML20:RWK20,ICML20:ZXHNCSK20,ECCV20:AMH20}.
The majority of established attacks are \emph{untargeted}, aiming to avoid
the correct classification~\cite{icml20:autopgd, iclr18:PGD,
  iclr15:fgsm, iclr17:i-fgsm, iclr14:lbfgs, cvpr16:deepfool,
  cvpr18:momentum, Euro16:JSMA, ICMLA17:semantic, nips19:sparse,
  ICML18:spsa, ccs16:eyeglasses, Oakland17:CarliniWagner,
  iclr16:featureadv, Oakland20:HopSkip, ICML18:noise}, while some
previous works explore \emph{targeted} adversarial attacks,
aiming to cause an input to be misclassified as a member of a \emph{single} specific,
incorrect class~\cite{icml20:autopgd,
  ICML22CGD, ccs16:eyeglasses, Oakland17:CarliniWagner}.
\emph{Robustness} is defined as a model's ability to resist evasion
attacks.  One common method to assess untargeted robustness is to
measure the model's accuracy on evasive
examples~\cite{arxiv21:robustbench, icml20:autopgd}.
Targeted robustness can be assessed by the model's resistance
to predict target classes chosen uniformly at random~\cite{ICML22CGD}.

\subsection{Motivation}
\label{sec:metric:motivation}

We suggest that untargeted and targeted robustness, as defined, are not sufficient
to accurately assess risk
for many real-world attack scenarios:
such attack scenarios could be complicated and involve more than
one misclassification, as in the traffic sign scenario discussed
in \secref{sec:intro}.
In this scenario, attackers might
perturb speed limit and delimit signs  
(signs that restrict the speed limit to specific values or mark the end
of previous such restrictions; see \figref{fig:signs}) into signs that
 require an immediate stop, including stop signs, no-entry signs, and no-vehicle
signs, or
 display a limit much lower than the actual limit, such as no more than half of the actual limit.

As another example, suppose students in a class are trying to access
materials (e.g., gradebooks) normally accessible only to TAs and professors.
The students might succeed by impersonating \emph{any} of the TAs or
professors, even if they cannot impersonate a specific TA or professor.
However, the students cannot succeed if they only impersonate other students.
The untargeted or targeted setting is not sufficient for this scenario.

The burglary example described in \secref{sec:intro}
serves as a more complicated example of an attack scenario.
Access to the vault
 is mediated by facial recognition and is granted only if \emph{several} of the staff are recognized as trying to open the vault together. 
The burglars thus succeed only if they are able to impersonate \emph{several distinctive members} of
the staff.
Which burglars (from a larger group) will attempt impersonation, and
which staff the impersonations will target (from among those who have
access to the vault), is at the burglars' discretion. 

Neither targeted nor
untargeted robustness intuitively corresponds well to either of these
 attack scenarios, as they measure the likelihood of
 successfully inducing \emph{any} misclassification or a misclassification to a
 \emph{single, specific} class. Neither of those corresponds to the
 attackers' goals in these scenarios, and, thus, 
 we need a different metric to better assess the risk.

\subsection{Definition}
\label{sec:metric:definition}

We propose \gbtrfull as a new metric that can assess
 the risk in the scenarios described in
\secref{sec:metric:motivation}. In prior work, individual evasion attacks
primarily sought to optimize
(mis)classification toward a specific class (targeted) or
optimize (mis)classification away from a specific class
(untargeted). The purpose of \gbtrfull{} is to formalize
group-based goals that encompass the results of multiple
evasion attacks, where ``groups'' consist of sets of classes.
We define \gbtrfull using an experiment, inspired by
experiments used to define cryptographic security
properties.  The experiment is parameterized by the
following:
\begin{itemize}
\item A \textit{classifier} is a possibly randomized algorithm that
  takes as input an
  \textit{instance} \instance in \instanceUniverse and returns a \textit{class}
  in \classUniverse.  That is, \instanceUniverse denotes the set of
  possible inputs to be classified and \classUniverse denotes the
  set of classes.  The experiment includes two classifiers:
  \begin{itemize}
  \item The \textit{ground truth classifier} \groundTruth is a
    deterministic classifier, i.e., a function.
  \item The \textit{targeted classifier} \classifier is a randomized
    or deterministic classifier.    
    Intuitively, the adversary will seek
    to mislead this classifier, i.e., to induce classifications by
    \classifier that differ from those by \groundTruth.
  \end{itemize}
\item Let $\impersonationGoalSet =
  \{\impersonationGoal{\impersonationGoalIdx}\}_{\impersonationGoalIdx=1}^{\impersonationGoalNmbr}$
  be a set of relations on \classUniverse, i.e., each
  $\impersonationGoal{\impersonationGoalIdx} \subseteq \classUniverse
  \times \classUniverse$.  Informally, the adversary's goal is to
  implement a misclassification defined by any
  $\impersonationGoal{\impersonationGoalIdx} \in
  \impersonationGoalSet$.  That is, in order to ``win,'' the adversary must,
  for some $\impersonationGoal{\impersonationGoalIdx} \in
  \impersonationGoalSet$, achieve \textit{every} misclassification
  $(\sourceclass, \targetclass) \in
  \impersonationGoal{\impersonationGoalIdx}$, in the sense that the
  adversary successfully transforms a given instance \instance of
  class \sourceclass (according to \groundTruth, $\groundTruth(\instance)=\sourceclass$) into an instance
  \instanceAlt that is classified in \targetclass by \classifier, while 
  $\groundTruth(\instanceAlt)=\sourceclass$.
It is worth pointing out that $\classifier(\instance)$ might be different from $\groundTruth(\instance)$, and with specific choices of $\impersonationGoalSet$, \instanceAlt might be the same as \instance.

\item \allowedPert is a predicate indicating whether \instanceAlt
  is ``close enough'' to \instance, e.g., according to some distance
  metric.  For the adversary to ``win,'' the instance \instanceAlt
  generated by modifying \instance must also satisfy
  $\allowedPert(\instance, \instanceAlt) = 1$.
  
\item \instanceSetGen is an algorithm generating instances
  $\instanceSet \subseteq \instanceUniverse$ by sampling them from
  some distribution.  
  Conventionally, $\vert \instanceSet \vert=1$. In \secref{sec:metric:space}, we will explain that certain choices of $\impersonationGoalSet$ are only achievable when  $\vert \instanceSet \vert>1$ while these choices correspond to realistic attack scenarios.
  Informally, \instanceSetGen is the environment
  that produces instances for which the adversary can try to induce
  misclassifications.

\item The adversary \adversary is an algorithm that takes as input a
  set $\instanceSet \subseteq \instanceUniverse$ and produces
  $\perturbations \subseteq \instanceSet \times \instanceUniverse$.
  Informally, if $(\instance, \instanceAlt) \in \perturbations$, then
  the adversary changes \instance into \instanceAlt to satisfy the
  properties above.  The adversary also has white-box access to
  \allowedPert, \groundTruth, \classifier, \instanceSetGen, and
  \impersonationGoalSet, which are public parameters of the
  experiment.

\end{itemize}

The formal definition of the experiment in which the adversary
\adversary participates is as follows:

\begin{center}
\begin{minipage}{3.5in}
\begin{tabbing}
*** \= *** \= \kill
Experiment $\experiment{\impRel}{\allowedPert,\groundTruth,\classifier,\instanceSetGen,\impersonationGoalSet}(\adversary)$ \\
\> $\instanceSet \gets \instanceSetGen()$ \\
\> $\perturbations \gets \adversary(\instanceSet)$ \\
\> $\impersonationSet \gets \{(\groundTruth(\instance), \classifier(\instanceAlt)) : (\instance, \instanceAlt) \in \perturbations\}$ \\
\> if $\left(\impersonationSet \in \impersonationGoalSet \wedge \forall (\instance, \instanceAlt) \in \perturbations: (\instance \in \instanceSet \wedge \allowedPert(\instance, \instanceAlt) = 1)\right)$ \\
\>\> return 1 \\
\> else \\
\>\> return 0
\end{tabbing}
\end{minipage}
\end{center}
\noindent
The adversary \adversary is run on the input instances $\instanceSet$ generated by $\instanceSetGen$. The relation achieved $\impersonationSet$ is computed based on the results of the adversary  \adversary, the ground truth classifier \groundTruth, and the targeted classifier \classifier. The experiment returns 1 (i.e., the adversary succeeds) if  \allowedPert indicates that the perturbation is small enough according to some distance metric \emph{and} $\impersonationSet$ matches some $\impersonationGoal{\impersonationGoalIdx}$ in the desired set $ \impersonationGoalSet$. The experiment returns 0 otherwise.

We define the \impRel advantage of \adversary to be
\[
\advantage{\impRel}{\allowedPert,\groundTruth,\classifier,\instanceSetGen,\impersonationGoalSet}(\adversary) = \prob{\experiment{\impRel}{\allowedPert,\groundTruth,\classifier,\instanceSetGen,\impersonationGoalSet}(\adversary) = 1}
\]
where the probability is taken w.r.t random choices made by
\instanceSetGen, \classifier, and \adversary.  Analogously, the
\gbtrfull is
\[
\robustness{\allowedPert,\groundTruth,\classifier,\instanceSetGen,\impersonationGoalSet}(\adversary) = \prob{\experiment{\impRel}{\allowedPert,\groundTruth,\classifier,\instanceSetGen,\impersonationGoalSet}(\adversary) = 0}
\]
  
By requiring properties of each
$\impersonationGoal{\impersonationGoalIdx} \in \impersonationGoalSet$,
we can express cases of interest.  For example, by requiring each
\impersonationGoal{\impersonationGoalIdx} to be a function, we require
that input instances \instance in distinct classes each be used to
impersonate only one class.

\subsection{A Broader Attack Space}
\label{sec:metric:space}

The definition of \gbtafull sheds light on a broader space of attacks
than had previously been explored with attacks or defenses.
As we introduced in \secref{sec:metric:motivation}, established
attacks are either untargeted, so that
$\impersonationGoalSet= \bigcup_{\sourceclass \in
  \classUniverse} \bigcup_{\targetclass \in
  \classUniverse\setminus\{\sourceclass\}} \left\{\{(\sourceclass,
\targetclass)\}\right\}$, avoiding correct classifications;
or targeted so that $\impersonationGoalSet=
\bigcup_{\sourceclass \in \classUniverse} \left\{\{(\sourceclass,
\targetclass{\sourceclass})\}\right\}$ for a specific
$\targetclass{\sourceclass} \in
\classUniverse\setminus\{\sourceclass\}$, seeking a specific impersonation.  
For example, untargeted attacks against GTSRB would be represented
as the former: the set of sign types would be represented by
$\classUniverse$, and attackers would attempt to
perturb a sign $\instance$ so that instead of being correctly classified as
belonging to class $\sourceclass$ it is misclassified as any other class
$\targetclass \in \classUniverse\setminus\{\sourceclass\}$.

To our knowledge, no existing evasion attack uses choices of
$\impersonationGoalSet$ other than the two listed above.
Consistently with that, we were unable to find defenses that are designed specifically for choices of $\impersonationGoalSet$ other than the two.
However, as our example scenarios start to illustrate, there are many
more other choices of $\impersonationGoalSet$ worthy of examination.
In the example where students are trying to access restricted materials,
\sourceclass could be any one of the students and \targetclass could be any one of the TAs or professors.
We denote the set of all student classes as $\sourceclassset$ and the
set of TAs and professors as $\targetclassset$, where $\sourceclassset
\subseteq \classUniverse$, $\targetclassset \subseteq
\classUniverse$, and $\sourceclassset$ is disjoint from $\targetclassset$. 
An attack succeeds if any student can impersonate any one of the TAs or professors, and thus we have 
$\impersonationGoalSet =
\bigcup_{\sourceclass \in \sourceclassset} \bigcup_{\targetclass \in
  \targetclassset} \{\{(\sourceclass, \targetclass)\}\}$, which is
different from the $\impersonationGoalSet$ in traditional targeted or
untargeted attacks.

In the example where attackers are perturbing traffic signs to slow down traffic,  
$\sourceclass$ could be any of the speed limit and delimit signs. The set of speed limit and delimit signs is now  $\sourceclassset \subseteq \classUniverse$.
However, for each $\sourceclass \in \sourceclassset$, the set of classes the adversary wishes to target might be different.
For example, a 20~KPH sign might be perturbed as a stop, no-entry, or
no-vehicle sign, whereas a 120~KPH sign might be perturbed as a 20,
30, 50, 60~KPH, stop, no-entry, or no-vehicle sign. 
Thus, attackers might have \emph{different sets of target classes
$\targetclassset_{\sourceclass}$ for different choices of
$\sourceclass$}. 
Attackers succeed in slowing traffic down if they perturb any of the
speed limit and delimit signs into any of the corresponding target
classes, and thus we have 
$\impersonationGoalSet =
\bigcup_{\sourceclass \in \sourceclassset} \bigcup_{\targetclass \in
  \targetclassset_{\sourceclass}} \{\{(\sourceclass,
\targetclass)\}\}$.

These new choices of $\impersonationGoalSet$ are able to describe the goal of
attackers trying to slow traffic down and students aiming to steal
access-restricted materials, while traditional $\impersonationGoalSet$
of untargeted or targeted attacks are not able to do so.  Formalizing
the attackers' goals in this way reveals that current evasion attacks
are not optimized for those goals.

Notice also that established attacks
count success on a
per-input-instance basis, using $\vert \instanceSet \vert=1$ where 
$\instanceSet$ is the set of input instances sampled at a time,
although $\instanceSet$ might be resampled and $ \experiment{\impRel}{\allowedPert,\groundTruth,\classifier,\instanceSetGen,\impersonationGoalSet}(\adversary)$
might be repeated many times.
Here we
examine cases where $\vert \instanceSet \vert>1$,
 enabling
consideration of attacker goals for which, for example, each
$\impersonationGoal{\impersonationGoalIdx} \in \impersonationGoalSet$
is a surjective function mapping classes $\sourceclassset$ to a target
set $\targetclassset$ of classes where $\sourceclassset \cap
\targetclassset = \emptyset$.  
In the burglary example, in which
 burglars 
impersonate several staff members of a bank to hack into a vault,
$\instanceSet$ is a set of images of the burglars at the time of the attack.
Different burglars might impersonate different staff 
and hence $\vert \instanceSet \vert>1$.
For each $\impersonationGoal{\impersonationGoalIdx}$, 
$\sourceclassset$ is a set consisting of a subset of burglars (since a subset may be enough to impersonate a sufficient number of bank staff) and
$\targetclassset$ is
a set of several staff who together are allowed access to the vault. 
To achieve an $\impersonationGoal{\impersonationGoalIdx}$, burglars might need to use more than one 
$\instance \in \instanceSet$.
In this attack scenario, 
compared with $\impersonationGoalSet$ of untargeted or targeted attacks,
the new choice of $\impersonationGoalSet$ also intuitively better
depicts the burglars' goal: successfully impersonating multiple different people.
 We propose three attack strategies \adversary in
\secref{sec:attacks:strategy} that boost
$\advantage{\impRel}{\allowedPert,\groundTruth,\classifier,\instanceSetGen,\impersonationGoalSet}(\adversary)$
for this new $\impersonationGoalSet$.

Our work serves as a step to search a wider space outside the crowded paradigm
of existing works.  New choices of $\impersonationGoalSet$
depict attack scenarios that often-used choices cannot.

\subsection{Experiment Setup}
\label{sec:metric:setup}

Now we turn to the experiment setups we employed to corroborate that \gbtrfull  complements our understanding of robustness
from previously established metrics.

\subsubsection{Threat Model}

Adversaries can create more successful
perturbations by acquiring more information about the architectures
and weights of models~\cite{CVPR18:Greybox}. The most successful
attacks are white-box attacks, where the adversaries have access to all
weights of models~\cite{NeurIPS20:Initializations,Usenix22:DosDonts}.
Accordingly, we use white-box 
evasion attacks to evaluate models to better understand the worst-case
threat and to more accurately
evaluate the existing defenses in the presence of the strongest
adversaries.

\subsubsection{Datasets}

We used three image datasets and one text dataset to empirically measure $\advantage{\impRel}{\allowedPert,\groundTruth,\classifier,\instanceSetGen,\impersonationGoalSet}(\adversary)$ 
and $\robustness{\allowedPert,\groundTruth,\classifier,\instanceSetGen,\impersonationGoalSet}(\adversary)$
in the three scenarios described in \secref{sec:metric:motivation}:
 We used a traffic-sign dataset (\gtsrb)~\cite{IJCNN13:GTSRB} for the scenario where attackers are trying to perturb traffic signs.  \gtsrb consists of images of 43 traffic signs, which include but are not limited to the speed limit, speed delimit, and signs that require an immediate stop (see \figref{fig:signs}).
In scenarios where students are trying to steal access-restricted materials and burglars are trying to hack a bank, one group of attackers--students or burglars, respectively--is trying to impersonate a group of victims.
In both scenarios, the attackers and victims are two mutually exclusive groups of people.
We used a human-face dataset (\pubfig)~\cite{ICCV09:Pubfig}, previously used in adversarial machine learning
  studies~\cite{ccs16:eyeglasses, tops19:GeneralFramework,
    iclr20:Wu2020Defending}, for both scenarios.
\pubfig consists of images of 60 identities and an average of 128 images per identity.
Face-recognition \dnn{}s may need to classify more than 60 identities; some of these identities might be neither attackers nor victims of impersonation.
However, 
the existing defense only used 10 identities~\cite{iclr20:Wu2020Defending}.
With more than 60 identities,
we could neither find a benchmark that has performance close to the existing defense~\cite{iclr20:Wu2020Defending} nor could we train one.
As an alternative, we used an object-recognition dataset
(\imagenet)~\cite{cvpr09imagenet} to mimic scenarios where there exist many identities that are neither attackers nor victims. 
\imagenet consists of images of 1000 objects, and we use these 1000 object classes to mimic 1000 identities.
Besides image datasets, we also used one text dataset, \sst~\cite{EMNLP13:SST5}. \sst has five classes, namely ``very positive", ``positive", ``neutral", ``negative" and ``very negative". 

\subsubsection{Benchmarks}
White-box evasion attacks have proved successful nearly $100\%$ of the time on models not specially tuned to be robust, but these attacks are less effective against
models that have been tuned to be robust~\cite{iclr18:PGD,NeurIPS19:FreeTraining}. 
Thus, to fairly compare the effectiveness of attacks, and to precisely compute $\advantage{\impRel}{\allowedPert,\groundTruth,\classifier,\instanceSetGen,\impersonationGoalSet}(\adversary)$ and $\robustness{\allowedPert,\groundTruth,\classifier,\instanceSetGen,\impersonationGoalSet}(\adversary)$ in the scenarios described in \secref{sec:metric:motivation}, we ran attacks against defended models.
In particular, we used the following state-of-the-art defenses against white-box evasion attacks
as benchmarks.

\subsubsection*{\gtsrb{}}
For \gtsrb{}, we trained defenses using the free adversarial
  training algorithm~\cite{NeurIPS19:FreeTraining}. We used two
  different $L_p$-norms, $L_{\infty}=8/255$ and $L_2=0.5$, as previous works did for images with similar sizes~\cite{icml20:autopgd,ICML22CGD}.
  We used six
  different architectures including five established architectures: 
VGG~\cite{iclr15:VGG},
 ResNet~\cite{cvpr16:ResNet}, SqueezeNet~\cite{arxiv16:SqueezeNet},
 ShuffleNet~\cite{ECCV18:ShuffleNet}, and
 MobileNet~\cite{cvpr18:MobileNet}.
 For each combination of architecture and $L_p$-norm, we trained 100
 instances for 100 iterations using the same implementation and
 data. The order of samples was also the same while training model instances. 
   The only
 difference between instances of the same architecture
 was the random
 initialization of weights.

 \subsubsection*{\pubfig{}}
  For the \pubfig{}
  dataset~\cite{ICCV09:Pubfig}, 
  in line with
  prior work, we preprocessed the face images by taking central
  crops and aligning faces to frontal poses via affine
  transformations~\cite{ccs16:eyeglasses,tops19:GeneralFramework,
    SPL16:MTCNN,cvpr15:facenet}. We split the data into 
  70\%-20\%-10\% for training, testing, and validation,
  respectively.
    We adversarially trained \dnn{}s via Wu et al.'s
    method~\cite{iclr20:Wu2020Defending}, using their
    implementation. Starting from a pre-trained feature extractor
    based on the VGG architecture~\cite{iclr15:VGG}, Wu et al.\ attached a two-layer
    classification head and conventionally trained the \dnn{},
    minimizing cross entropy. Next, they fine-tuned their model over
    several epochs of adversarial training. In each iteration of
    adversarial training, their method located the central region of an
    adversarial rectangular patch via a gradient-based
    search. Subsequently, the rectangular patch was perturbed 
    to induce misclassification.
    The resulting misclassified 
    image with the patch and a correct label was then used to update
    the \dnn{}'s weights. We trained the \dnn{} conventionally for 30
    epochs, and adversarially for 5 epochs, using Wu et al.'s
    default choice of the optimizer (Adam~\cite{iclr15:Adam}), step size (4),
    and batch size (64 for conventional training, and 32 for
    adversarial training). 
    One common attack used to evaluate the robustness of facial recognition systems is the eyeglasses attack~\cite{ccs16:eyeglasses} 
    which limits the perturbation to be within an eyeglasses-frame-shaped region (rather than any $L_p$ distance),
    simulating that attackers wear carefully painted eyeglasses to evade face recognition.
    The adversarially trained \dnn{} achieved
    98.25\% benign test accuracy and 45.43\% untargeted robustness against
    the eyeglass attack. By contrast, the conventionally trained model
    achieved 99.80\% benign accuracy but only 9.14\% untargeted robustness. As we described in \secref{sec:metric:space},
    untargeted robustness is $\robustness{\allowedPert,\groundTruth,\classifier,\instanceSetGen,\impersonationGoalSet}(\adversary)$
    when $\impersonationGoalSet= \bigcup_{\sourceclass \in
    \classUniverse} \bigcup_{\targetclass \in
    \classUniverse\setminus\{\sourceclass\}} \left\{\{(\sourceclass,
  \targetclass)\}\right\}$ .

  \subsubsection*{\imagenet}
Face-recognition \dnn{}s might need to classify more than 60 identities; many of these identities are neither attackers nor victims.
However, we were not able to find a pre-trained face-recognition defense that uses more than 60 identities. 
For example,
Wu et al.\ used a subset of VGG-Face~\cite{iclr15:VGG} which includes ten identities~\cite{iclr20:Wu2020Defending}.
We also tried to train face-recognition defense using existing methods, but the performance of our trained instances was much worse than the performance (of defenses with less than 60 identities) reported in previous works.
As a mitigation, we used \imagenet instead. \imagenet has 1000 classes and
 we found two pre-trained state-of-the-art defenses on \imagenet by Salman et
  al.~\cite{NeurIPS20:robustness}. One of them was trained with 
  adversarial perturbations of $L_{\infty}$-norm of 8/255, and the
  other one was trained perturbations of $L_2$-norm  of 3.0.
  We used these two defenses to mimic face-recognition defenses that are capable of recognizing 1000 identities.
  
  \subsubsection*{\sst}
  We used a pre-trained model~\cite{UNSO} on \sst, which achieves $55.8\%$ accuracy (within top five performance at the time we conducted experiments, according to a leaderboard~\cite{leaderborad:SST5}). This model has not been specifically tuned for robustness.

\subsubsection{Measurement Process}
\label{sec:metric:background:process}
We implemented \instanceSetGen 
  for each test $\impersonationGoalSet$ on different datasets. 
To measure $\robustness{\allowedPert,\groundTruth,\classifier,\instanceSetGen,\impersonationGoalSet}(\adversary)$,
 we still conventionally used $\vert \instanceSet \vert=1$ 
where $\instanceSetGen$ always outputs $\instanceSet =\{  \instance \}$,
where $\instance$ is one input instance, uniformly sampled from all
instances associated with some $\sourceclass \in \sourceclassset$.

On the \gtsrb{} dataset, we tried to perturb speed limit and delimits
signs to signs that would mandate (1) an immediate stop or (2) no more than half of the actual limit (shown in
\figref{fig:signs}). $\impersonationGoalSet =
\bigcup_{\sourceclass \in \sourceclassset} \bigcup_{\targetclass \in
  \targetclassset_{\sourceclass}} \{\{(\sourceclass, \targetclass)\}\}$.
  Images from different classes may not have the same set of
target classes. As we introduced in \secref{sec:metric:space},
for each $\sourceclass \in \sourceclassset$, the set of targeted
classes $\targetclassset_{\sourceclass}$ might be different. A 20 KPH
sign might be perturbed as a stop, no-entry, or no-vehicle sign, and a
120 KPH sign might be perturbed as a 20 KPH, 30 KPH, 50 KPH, 60 KPH, stop,
no-entry, or no-vehicle sign. 

On the \pubfig{} dataset, 
we have $\impersonationGoalSet =
\bigcup_{\sourceclass \in \sourceclassset} \bigcup_{\targetclass \in
  \targetclassset} \{\{(\sourceclass, \targetclass)\}\}$ as in the scenario where students are trying to steal access-restricted materials.
We randomly selected two mutually exclusive
sets of classes $\sourceclassset$ and $\targetclassset$, and tried to
perturb all images associated with $\sourceclassset$ as
$\targetclassset$. We used the following
sizes of $\sourceclassset$ and $\targetclassset$: 
\begin{align}
(\vert \sourceclassset \vert, \vert \targetclassset \vert) \in{} &  \{ (10,10), (10,20),(10,30),(10,40),(10,50), \nonumber\\
      &  (20,10),(20,20),(20,30),(20,40), (30,10), \nonumber\\
  & 30,20),(30,30),(40,10),(40,20),(50,10)  \}
  \label{eqn:sizes}
 \end{align}
These choices include all possible choices of $\vert \sourceclassset \vert$ or $\vert \targetclassset \vert$ that are multiples of 10.
For each $(\vert \sourceclassset \vert, \vert \targetclassset \vert)$,
we randomly selected 5 different pairs of subsets $\sourceclassset$ and
$\targetclassset$. 

On the \imagenet dataset, 
we still have $\impersonationGoalSet =
\bigcup_{\sourceclass \in \sourceclassset} \bigcup_{\targetclass \in
  \targetclassset} \{\{(\sourceclass, \targetclass)\}\}$,
  because the attack scenario is the same as on \pubfig{}.
We first randomly selected 60 classes, and
then selected $\sourceclassset$ and $\targetclassset$ sets of sizes in \eqnref{eqn:sizes}
from these
60 classes, as we did for \pubfig{}.

On the \sst dataset
we used four different goals of the adversary: 1) perturb positive instances as nonpositive (i.e. perturb instances from the "very positive" or "positive" classes as any of the rest three classes) 2) perturb negative instances as nonnegative 3) perturb each instance as a more positive class and 4) perturb each instance as a more negative class. 
For each of these four goals, similar to the traffic sign scenario, $\impersonationGoalSet =
\bigcup_{\sourceclass \in \sourceclassset} \bigcup_{\targetclass \in
  \targetclassset_{\sourceclass}} \{\{(\sourceclass,
\targetclass)\}\}$ with respect to different choices of $\sourceclassset$ and $\targetclassset_{\sourceclass}$.

On all the benchmarks trained with $L_p$-norm (benchmarks on \gtsrb and \imagenet), we ran
\autopgd~\cite{icml20:autopgd} attacks using the same $L_p$
distance.
That is, in
$\experiment{\impRel}{\allowedPert,\groundTruth,\classifier,\instanceSetGen,\impersonationGoalSet}(\adversary)$,
if attacking a model \classifier trained with $L_\infty =
8/255$, for example, then $\allowedPert(\instance, \instanceAlt) = 1$ if
and only if $L_\infty(\instance, \instanceAlt) \le 8/255$.
To the 
best of our knowledge, \autopgd attacks are the strongest currently
available $L_p$-norm attacks that do not require model-specific
tuning.
On the \doa defenses, we ran eyeglasses
attacks~\cite{ccs16:eyeglasses} implemented by the authors of
\doa~\cite{iclr20:Wu2020Defending}. 
On the \sst dataset, we ran \tpgd attacks \cite{ACL23:TPGD},
a state-of-the-art text domain attack that empirically achieves human imperceptibility.
In our experiments, we exclusively
modified the loss function of any attacks we used. Using their default settings, we ran \autopgd for
attacks for 100 iterations, eyeglasses attacks for 300 iterations, and \tpgd attacks for 100 iterations. 

\subsection{Results}
\label{sec:metric:results}
\begin{figure}[t!]
\centerline{\includegraphics[width=0.95\columnwidth]{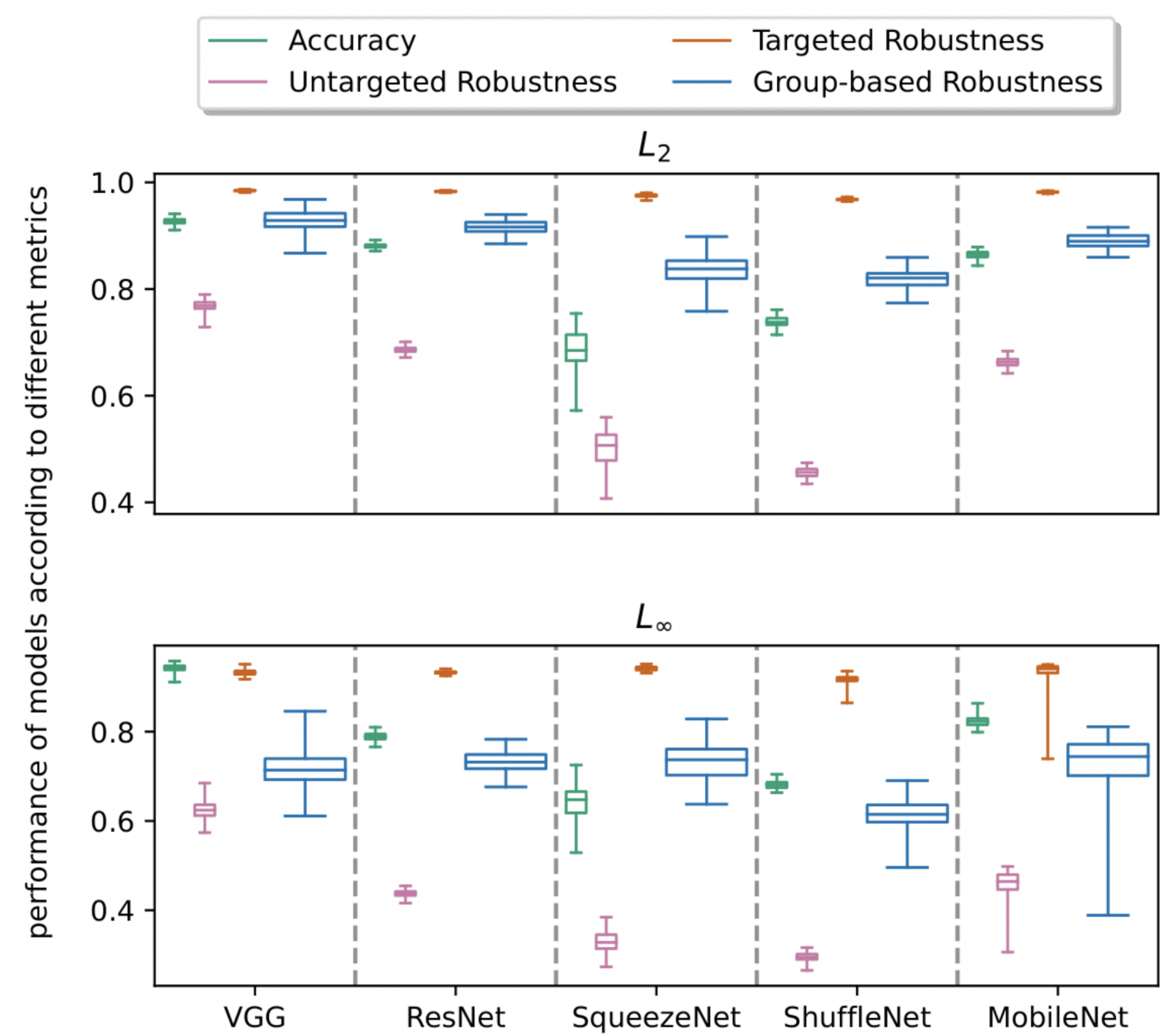}}
\caption{Performance of models on \gtsrb{} measured by four metrics: accuracy,
  untargeted robustness, targeted robustness, and \gbtrfull. With each
  combination of $L_p$-norm and architecture, the distribution of
  \gbtrfull, depicted as the wider boxes, is different from those of the other three
  metrics. With each combination, the performance of models varies only due to different randomly initialized weights, using seeds 0 -- 99.\keane{Is it possible to standardize the capitalization of the y-axis labels across figures? some are uncapatalized, some are not}}
\label{fig:metrices}
\end{figure}

\begin{figure}[t!]
\centerline{\includegraphics[width=0.95\columnwidth]{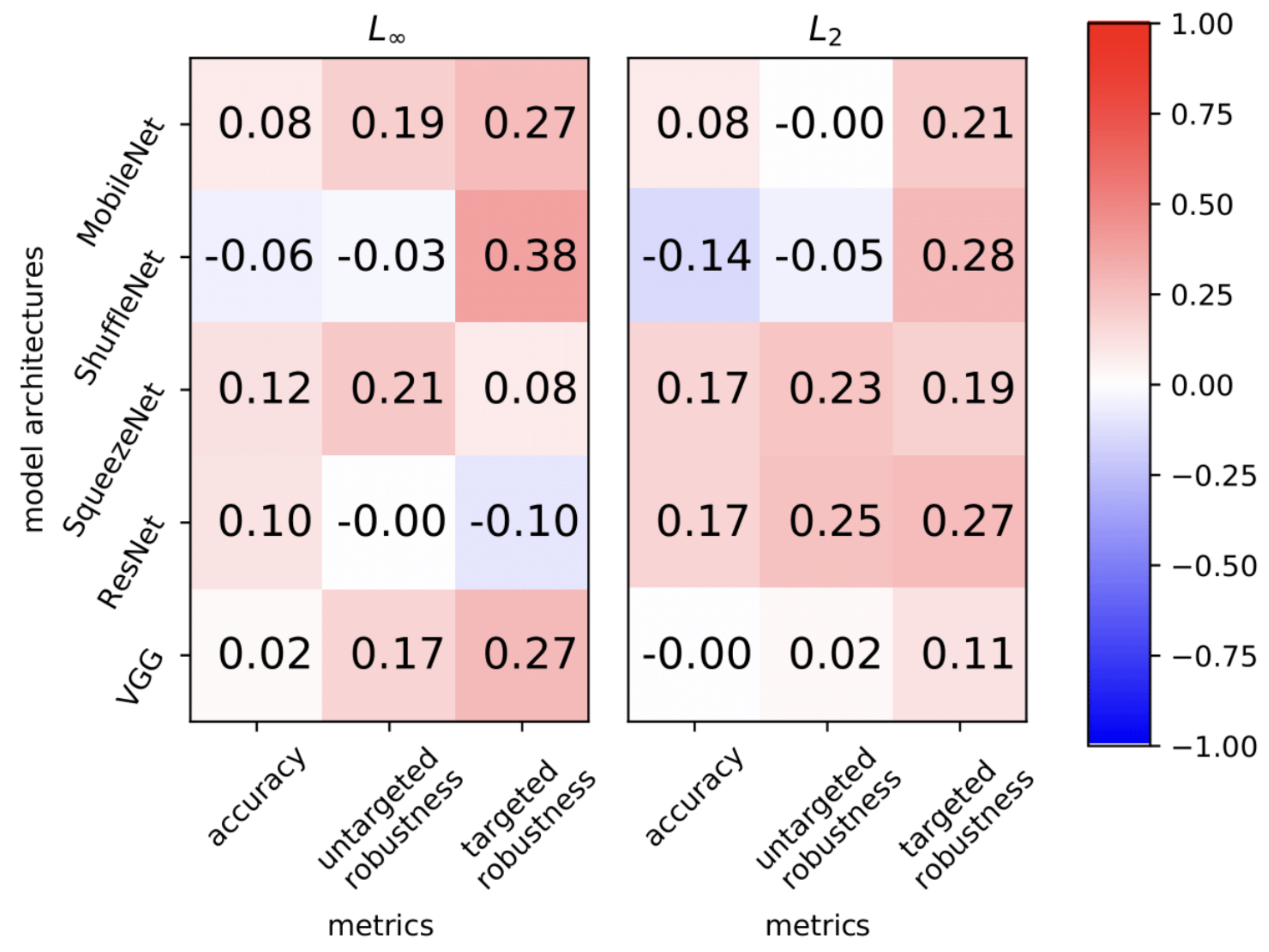}}
\caption{Pearson correlation coefficients between \gbtrfull and three
  existing metrics: accuracy, untargeted robustness, and targeted
  robustness on \gtsrb{}. Across most of the combinations of model architecture
  and $L_p$-norm, the correlations are negligible or weak as the
  coefficients have a magnitude smaller than
  0.4~\cite{AA18:correlation}.}
\label{fig:corr}
\end{figure}

 \secref{sec:metric:space} showed how our new metric, \gbtrfull,
conceptually accommodates real-world attack scenarios that existing metrics cannot.
\mahmood{it's weird to refer to section 2 from within section 2. the pointer should be more specific (point to a subsection) or potentially avoided. also, the sentence does not parse well. does this aim to say that the new metric can conceptually accommodate attack scenarios that other metric cannot?} \weiran{both addressed}
To
 empirically show this, 
we measured \gbtrfull along with three metrics:
accuracy, untargeted robustness, and targeted robustness.
As we described in \secref{sec:metric:space}, untargeted robustness is $\robustness{\allowedPert,\groundTruth,\classifier,\instanceSetGen,\impersonationGoalSet}(\adversary)$ when $\impersonationGoalSet= \bigcup_{\sourceclass \in
  \classUniverse} \bigcup_{\targetclass \in
  \classUniverse\setminus\{\sourceclass\}} \left\{\{(\sourceclass,
\targetclass)\}\right\}$, and targeted robustness is $\robustness{\allowedPert,\groundTruth,\classifier,\instanceSetGen,\impersonationGoalSet}(\adversary)$ when $\impersonationGoalSet=
\bigcup_{\sourceclass \in \classUniverse} \left\{\{(\sourceclass,
\targetclass{\sourceclass})\}\right\}$
for a specific
$\targetclass{\sourceclass} \in \classUniverse\setminus\{\sourceclass\}$.
When measuring \gbtrfull, we used choices of $\impersonationGoalSet$ motivated by attack scenarios described in \secref{sec:metric:motivation}.

On the
benchmarks we trained on \gtsrb{}, as shown
in \figref{fig:metrices}, \gbtrfull has different distribution (mean
and range) from
the other three metrics. For some combinations of architecture
and $L_p$-norms, the range of \gbtrfull, the distribution barely
overlaps with those of other metrics. 
Meanwhile, the variance for \gbtrfull is higher than the variance of other metrics at most of the combinations of architecture
and $L_p$-norm.
Models with close performance by other metrics can have very different performance by \gbtrfull.

We verified the difference statistically: as shown in
\figref{fig:corr}, at each combination of architecture and $L_p$-norm,
the Pearson correlation coefficients between \gbtrfull and each of the
three metrics are always between $-0.4$ to $0.4$.
According to Schober et al., correlations are weak or negligible of
the Pearson coefficient is 
between $-0.4$ and $0.4$~\cite{AA18:correlation}.
\Gbtrfull is always negligibly or weakly correlated with each of the
three metrics.

\begin{figure}[t!]
\centerline{\includegraphics[width=0.95\columnwidth]{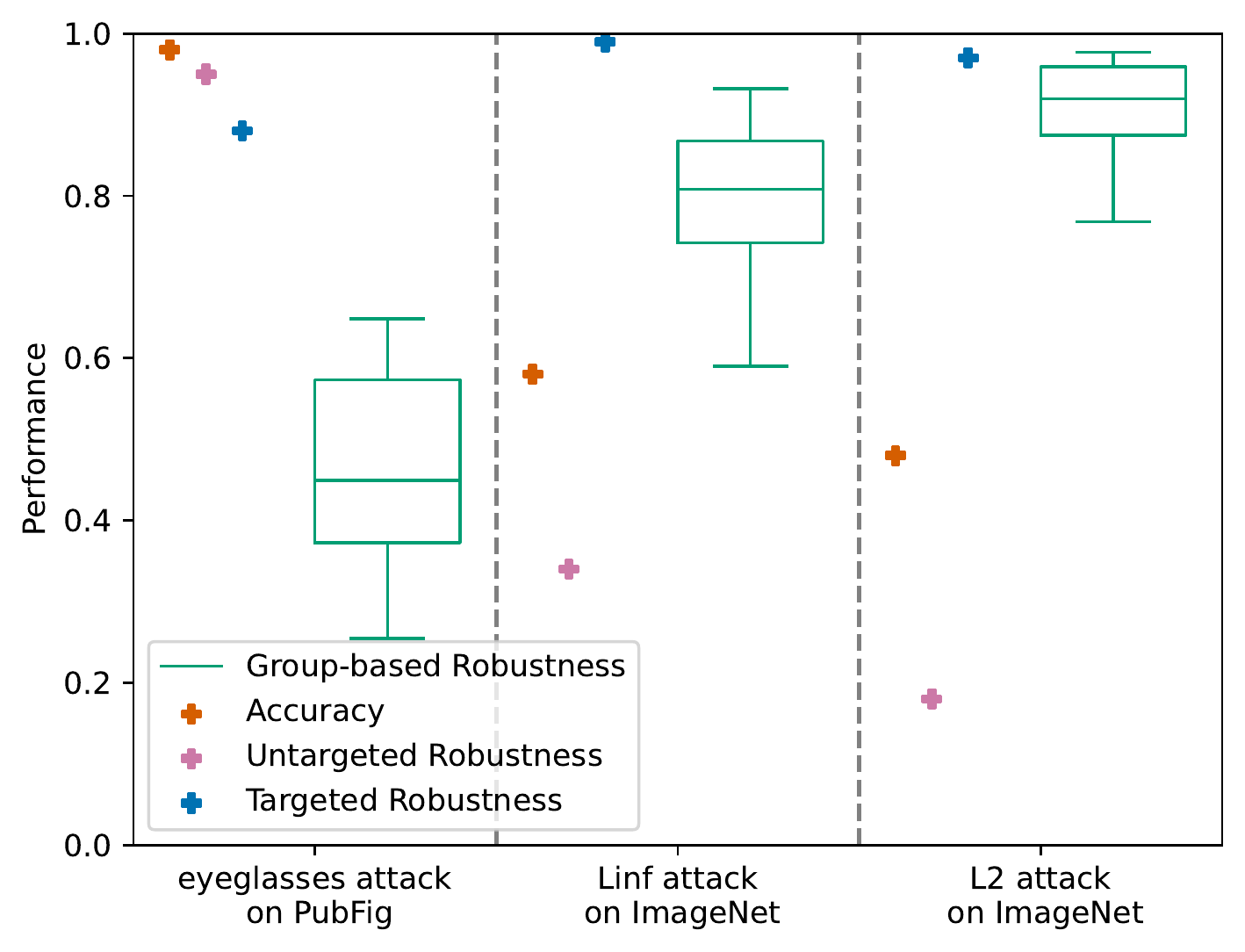}}
\caption{Performance of models measured by accuracy, untargeted
  robustness (UR), targeted robustness (TR), and
  \gbtrfull (\gbtrapprev). These models were adversarially trained with
  \pubfig{} and \imagenet. On
  each model, \gbtrapprev has a wide range due to different
  choices of $\targetclassset$ and $\sourceclassset$, whereas the
  other metrics report only a single value that is sometimes
  out of the \gbtrapprev range.}
%
%
%
%
\label{tab:metric}
\end{figure}

\begin{figure}[t!]
\centerline{\includegraphics[width=0.95\columnwidth]{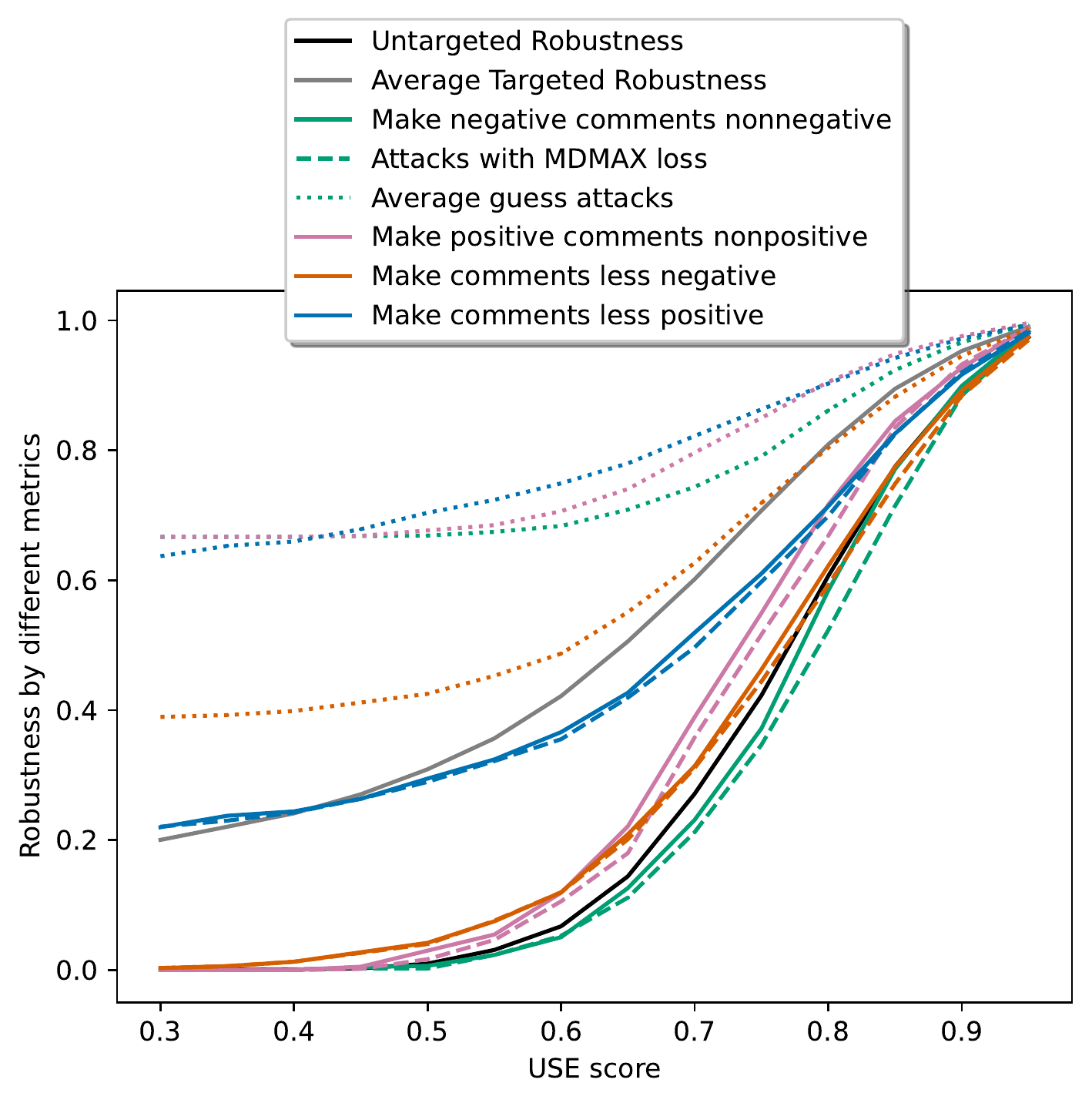}}
\caption{Robustness as defined by different metrics on the \sst dataset. The USE score, as used by T-PGD, denotes the imperceptibility of the perturbation: the higher the USE score is, the more imperceptible the perturbation is to humans. \Gbtrfull has a wide range, whereas untargeted robustness and targeted robustness are sometimes out of the range.}
\label{fig:textattacks}
\end{figure}

We also measured the performance of benchmarks we have on the \pubfig{} and
\imagenet datasets with these four metrics. As shown in
\tabref{tab:metric}, \gbtrfull reports a wide range on each of the
three models, whereas each of the other three metrics reports a single
number, some laying outside the range. \Gbtrfull
reports different numbers due to different choices of
$\sourceclassset$ and $\targetclassset$, which correspond to different 
attack scenarios, such as adversaries slowing traffic down or burglars
hacking into a vault at a bank. We also measured \gbtrfull, along with targeted and untargeted robustness as shown in \figref{fig:textattacks}. The benign accuracy is a constant number, and similar to what we saw on other datasets,  \gbtrfull reports a range, and targeted or untargeted robustness is sometimes outside the range.\keane{this and several other paragraphs, captions only have 1 or 2 words on the last line, which could look better if removed.}


\begin{myblock}{Takeaways (Metric)}
  \Gbtrfull $\robustness{\allowedPert,\groundTruth,\classifier,\instanceSetGen,\impersonationGoalSet}(\adversary)$ measures the robustness of models
  using different choices of $\impersonationGoalSet$ in accordance with
  the attack scenarios.
  Conventional metrics cannot measure the true threat in these
  sophisticated scenarios as accurately as \gbtrfull does. 
  Thus, we conclude that
  \gbtrfull offers a new meaningful
    assessment of model susceptibility to attacks in the real world
    compared to conventional metrics.
\end{myblock}

\section{More Efficient Attacks}
\label{sec:attacks}
In this section, we introduce several algorithms \adversary that either
increase the advantage of attacks
$\advantage{\impRel}{\allowedPert,\groundTruth,\classifier,\instanceSetGen,\impersonationGoalSet}(\adversary)$,
or boost the speed to achieve a close advantage of attacks,
helping attacks become more computationally efficient than existing na\"{i}ve attacks.
We start by describing two loss functions 
to help adversaries perturb one input instance so that the input instance is misclassified as any of a specific set of target classes
(\secref{sec:attacks:loss}). 
Then, we introduce three attack strategies 
to help adversaries perturb several input instances so that they are
misclassified as 
different target classes among a specific set
(\secref{sec:attacks:strategy}).

Later, we also leverage these new attacks to build defenses against
\gbtafull (see \secref{sec:defenses}).


\subsection {Attack Loss Functions}
\label{sec:attacks:loss}

In certain scenarios, adversaries may have a limited amount of time or
attempts to attack systems. In the bank robbery example, burglars
might only have a brief time window to access the face recognition
system, and they might trigger an alarm if they consecutively make many
failed attempts to impersonate bank staff. Meanwhile,
it might be impractically costly both for the attackers to try impersonating each of the employees as a brute-force approach,
 and for the bank to assess the \gbtrfull in such a manner, as the bank might have many staff (e.g., Dresdner Bank, a major European bank, has 50,659 employees~\cite{SACMAT01:Bank}). 
 Thus, we need \gbtafull that are more time-efficient than the
 brute-force approach. 
We designed two new loss functions that formalize the attackers' goal in such scenarios.

We came up with these two loss functions 
by modifying a state-of-the-art loss function for targeted attacks,
the Minimal Difference (\md) loss~\cite{ICML22CGD}, 
which outperforms well-known loss functions such as the Carlini-Wagner (CW) loss~\cite{Oakland17:CarliniWagner} and the Difference
of Logits Ratio (DLR) loss~\cite{icml20:autopgd}.
It aims to assign the highest logit to the target class:
\begin{equation}
\ell_{\md}=\sum_{i}ReLU(Z_i+\delta-Z_{\targetclass})
\end{equation}
where \textit{i}
iterates over all classes, $\targetclass$ is the target class, \textit{Z} is the logit, $\delta$ is a minimal value set to 
1e-15,
and \textit{ReLU} is the rectified linear unit
function.
Said differently, attacks \adversary with the \md loss have higher $\advantage{\impRel}{\allowedPert,\groundTruth,\classifier,\instanceSetGen,\impersonationGoalSet}(\adversary)$
than attacks with other previously proposed loss functions,
for targeted attacks with $\impersonationGoalSet=
\bigcup_{\sourceclass \in \classUniverse\setminus\{\targetclass\}} \left\{\{(\sourceclass,
\targetclass{})\}\right\}$ for a specific target class
$\targetclass \in \classUniverse$, perturbing more input instances
$\instance \in \instanceUniverse$ to be misclassified as the target
class $\targetclass$. 
The attack succeeds
  \emph{if and only if} the MD loss is zero.
When the \md loss is zero, $Z_{\targetclass}$ is larger than any other
$Z_i$ by at least $\delta$, and thus the attack succeeds. On the other hand,
if the attack succeeds, $Z_{\targetclass}$ is the largest logit, larger than any other $Z_i$ by at least the minimal value $\delta$; thus the \md loss is zero.

As we explained in \secref{sec:metric}, an adversary could be
interested in some set of target classes $\targetclassset$, rather than a
single target  
class $\targetclass$. In this case, an attack is considered successful if the adversary can cause an input from a class in \sourceclassset to be misclassified as \emph{any} class
within $\targetclassset$. 
More formally,
as we described in \secref{sec:metric:space},
in the example where students are trying to steal access-restricted materials, 
$\impersonationGoalSet =
\bigcup_{\sourceclass \in \sourceclassset} \bigcup_{\targetclass \in
  \targetclassset} \{\{(\sourceclass, \targetclass)\}\}$, where $\sourceclassset$ is the set of all students and $\targetclassset$ is the set of all TAs and professors. 
If a student impersonates \emph{any} of the TAs or professors, 
the attack succeeds.
In the example where attackers perturb signs to slow traffic down,
$\impersonationGoalSet =
\bigcup_{\sourceclass \in \sourceclassset} \bigcup_{\targetclass \in
  \targetclassset_{\sourceclass}} \{\{(\sourceclass, \targetclass)\}\}$,
  $\sourceclass$ is one of the speed limit and delimit signs,
and $\targetclassset_{\sourceclass}$ is the corresponding set of signs that can mislead traffic to be slower than intended.
  The attackers succeed if they can perturb a sign from the class
  $\sourceclass$ to be classified as any sign type in $\targetclassset_{\sourceclass}$.
  Notice that $\targetclassset_{\sourceclass}$ depends on
  $\sourceclass$, e.g., the attack succeeds if a 20~KPH sign is perturbed into a stop,
  no-entry, or no-vehicle sign; or if a 120~KPH sign is
  perturbed into a 20, 30, 50, 60~KPH, stop, no-entry, or no-vehicle
  sign.

A na\"{i}ve exhaustive approach to carry out such misclassification would be to launch $\vert
\targetclassset \vert$ targeted attacks, one for each $\target \in
\targetclassset$.  That is, to maximize $\advantage{\impRel}{\allowedPert,\groundTruth,\classifier,\instanceSetGen,\impersonationGoalSet}(\adversary)$,
\adversary invokes subroutines $\adversary_{\target}$ to find
an impersonation for each $\target \in \targetclassset$ independently.
\adversary succeeds when \emph{any} of the $\vert \targetclassset
\vert$ targeted attacks succeed.
When $\impersonationGoalSet= \bigcup_{\sourceclass \in
  \classUniverse} \bigcup_{\targetclass \in
  \classUniverse\setminus\{\sourceclass\}} \left\{\{(\sourceclass,
\targetclass)\}\right\}$ and so $\targetclassset =
\classUniverse\setminus\{\sourceclass\}$, this na\"{i}ve approach
finds more adversarial examples, obtaining higher
$\advantage{\impRel}{\allowedPert,\groundTruth,\classifier,\instanceSetGen,\impersonationGoalSet}(\adversary)$
than any other untargeted attacks \adversary that aim to directly
maximize
$\advantage{\impRel}{\allowedPert,\groundTruth,\classifier,\instanceSetGen,\impersonationGoalSet}(\adversary)$~\cite{icml20:autopgd}.
However, this na\"{i}ve exhaustive approach requires running targeted
attacks $\vert \targetclassset \vert$ times and thus is $\vert
\targetclassset \vert$ times more costly to run than untargeted
attacks. Another na\"{i}ve approach that does not suffer from the same
overhead is to randomly pick a class
$\targetclass \in \targetclassset$
and launch a targeted attack $\adversary_{\target}$
targeting \target.  However, we found this
approach finds significantly fewer adversarial examples, obtaining much smaller 
$\advantage{\impRel}{\allowedPert,\groundTruth,\classifier,\instanceSetGen,\impersonationGoalSet}(\adversary)$
(more details can be found  in \secref{sec:attack:loss:results}).

To address the shortcomings of the na\"{i}ve approaches,
we propose two new loss functions for
when $\impersonationGoalSet =
\bigcup_{\sourceclass \in \sourceclassset} \bigcup_{\targetclass \in
  \targetclassset} \{\{(\sourceclass, \targetclass)\}\}$ or $\impersonationGoalSet =
\bigcup_{\sourceclass \in \sourceclassset} \bigcup_{\targetclass \in
  \targetclassset_{\sourceclass}} \{\{(\sourceclass, \targetclass)\}\}$. These loss functions
 help attackers obtain larger 
  $\advantage{\impRel}{\allowedPert,\groundTruth,\classifier,\instanceSetGen,\impersonationGoalSet}(\adversary)$ 
  than the non-exhaustive na\"{i}ve approach,
  and obtain a close $\advantage{\impRel}{\allowedPert,\groundTruth,\classifier,\instanceSetGen,\impersonationGoalSet}(\adversary)$
  while consuming much less computation time than the exhaustive na\"{i}ve approach.

\subsubsection{The \mdgroup Loss} 
We propose a
new loss function following the intuition that 
attackers only need one class in the targeted set $\targetclassset$ to have a higher logit than any classes not in the set. 
In particular, we formalize the attackers'
goal---to assign a higher logit to some $\targetclass \in \targetclassset$ than to any $i \notin \targetclassset$---as
$ \sum_{i \notin \targetclassset}ReLU(Z_i+\delta-Z_{\targetclass})$.
This
term
is always non-negative, 
and
it is zero if and only if the corresponding $\targetclass$ has higher logit than any $i \notin \targetclassset$.
To capture that an adversary only needs one $\targetclass \in
\targetclassset$ to have higher logit than any $i \notin
\targetclassset$, we can write $\prod_{\targetclass \in \targetclassset} \sum_{i \notin
\targetclassset}ReLU(Z_i+\delta-Z_{\targetclass})$, which, again,
evaluates to zero if and only if the attack succeeds.
Due to the finite arithmetic of Python, in which we implement these loss
functions, this product can be $\infty$ and yield undefined gradients.
As a remedy, we compute the natural logarithm of the product instead and come up with 
the Minimal Difference Multiplied (\mdgroup) 
loss: 
\begin{equation}
\ell_{\mdgroup}=\sum_{\targetclass \in \targetclassset} ln(\sum_{i \notin \targetclassset}ReLU(Z_i+\delta-Z_{\targetclass}))
\end{equation}
where $\targetclass$ iterates over all classes in $\targetclassset$
and \textit{i} iterates over all classes not in $\targetclassset$. 
We acknowledge that mathematically $ln(0)$ is undefined, while
Python computes $ln(0)$ as $-\infty$. Each 
natural logarithm result is $-\infty$ if and only if a successful
attack has been found: $Z_{\targetclass}$ is larger than all
$Z_i$, and the whole equation is $-\infty$ if and only if at least one of the
natural logarithm results is $-\infty$. Thus $\ell_{\mdgroup}$ is
$-\infty$ if and only if a successful attack has been
found.\mahmood{the repetition of ``if and only if'' in the last three
  sentences is confusing. these should be streamlined better.}\weiran{addressed}

\subsubsection{The \mdmax Loss} 
Attackers only need one class in the targeted set $\targetclassset$ to have a higher logit than any classes not in the set. One strategy to achieve this is to 
greedily keep trying to increase the
current maximum logit from among the targeted classes,
perturbing
inputs toward the target class that is most likely
to succeed. 
We formalize this approach as 
the Minimal Difference Maximum (\mdmax) loss:
\begin{equation}
\ell_{\mdmax}=\sum_{i \notin \targetclassset}ReLU(Z_i+\delta-\max_{\targetclass \in \targetclassset} Z_{\targetclass})
\end{equation}
where \textit{i} iterates over all classes $\notin \targetclassset$ and the
largest $Z_{\targetclass}$ among all classes $\targetclass \in
\targetclassset$ is used. $\ell_{\mdmax}$ is also non-negative and zero
if and only if a successful attack has been found: if
$\ell_{\mdmax}>0$, there is at least one class $i \notin
\targetclassset$ that has higher logits than all classes $\targetclass
\in \targetclassset$; otherwise, if $\ell_{\mdmax}=0$, there is at least
one class $\targetclass \in \targetclassset$ that has higher logits
than all classes $i \notin \targetclassset$.

\subsubsection{Experiment Setup}
To illustrate the new loss functions' improvements for adversaries
\adversary, either in terms of advantage or speed, we used the same
threat model, datasets, and benchmarks as we did in
\secref{sec:metric:setup}. 
We also implemented \instanceSetGen 
  for each test $\impersonationGoalSet$ on different datasets as we did in \secref{sec:metric:setup}: 
we used conventionally $\vert \instanceSet \vert=1$ 
where $\instanceSetGen$ always outputs $\instanceSet =\{  \instance \}$,
and $\instance$ is one input instance, and uniformly sampled from all
instances associated with \emph{some} $\sourceclass \in
\sourceclassset$. In addition, we used the baselines and
measurement process below.
 
\subsubsection*{Baselines}
We created three baseline attacks to compare with attacks using loss functions proposed in \secref{sec:attacks:loss}.
As we elaborated in \secref{sec:metric},
adversaries could face scenarios where they succeed by forcing any
misclassifications within a set $\targetclassset$ of target classes. 
More formally, they seek to maximize $\advantage{\impRel}{\allowedPert,\groundTruth,\classifier,\instanceSetGen,\impersonationGoalSet}(\adversary)$, when 
$\impersonationGoalSet =
\bigcup_{\sourceclass \in \sourceclassset} \bigcup_{\targetclass \in
  \targetclassset} \{\{(\sourceclass, \targetclass)\}\}$  or
  $\impersonationGoalSet =
\bigcup_{\sourceclass \in \sourceclassset} \bigcup_{\targetclass \in
  \targetclassset_{\sourceclass}} \{\{(\sourceclass, \targetclass)\}\}$ (introduced in \secref{sec:metric:space}).
They could exhaustively iterate over all classes in
$ \targetclassset_{\sourceclass} $
or randomly pick a $\target \in  \targetclassset_{\sourceclass} $,
and launch targeted attacks. 
We depict these na\"{i}ve methods as follows: 
\begin{itemize}
\item The best guess: with every $\instance \in \instanceSet$ to be
  perturbed, \adversary could iterate over all $\target \in
  \targetclassset_{\sourceclass} $, and launch $\vert
  \targetclassset_{\sourceclass} \vert$ targeted attacks.  Each
  targeted attack $\adversary_{\target}$ aims to produce pairs
  $(\instance, \instanceAlt)$ such that $\classifier(\instanceAlt) \in
  \target$.  \adversary succeeds on this $\instance$ if \emph{any} of
  the $\vert \targetclassset_{\sourceclass} \vert$ targeted attacks
  succeed.  This approach tends to be the strongest na\"{i}ve
  approach, obtaining the highest
  $\advantage{\impRel}{\allowedPert,\groundTruth,\classifier,\instanceSetGen,\impersonationGoalSet}(\adversary)$.
  However, as it searches exhaustively, \adversary is $\vert
  \targetclassset_{\sourceclass} \vert$ times as expensive to run as
  each $\adversary_{\target}$.
\item The average guess: on each $\instance \in \instanceSet$,
  \adversary randomly picks a $\target$ from $\vert \targetclassset_{\sourceclass} \vert$  and
  outputs pairs $(\instance, \instanceAlt)$ such that
  $\classifier(\instanceAlt) = \target$.
  $\advantage{\impRel}{\allowedPert,\groundTruth,\classifier,\instanceSetGen,\impersonationGoalSet}(\adversary)$
  of the average guess attacks is the mean of all
  $\advantage{\impRel}{\allowedPert,\groundTruth,\classifier,\instanceSetGen,\impersonationGoalSet_{\target}}(\adversary)$
  whose $\impersonationGoalSet_{\target} = \bigcup_{\sourceclass \in
    \sourceclassset} \left\{\{(\sourceclass,
  \targetclass{})\}\right\},\target \in \targetclassset$.  The average
  running time of \adversary is the average running time of all
  $\adversary_{\target}$.

\end{itemize}


We run the two na\"{i}ve methods with $\ell_{\md}$, the
state-of-the-art loss function for targeted attacks (introduced in
\secref{sec:attacks:loss}).

\subsubsection*{Measurement Process}
\label{sec:attacks:loss:process}
We computed $\advantage{\impRel}{\allowedPert,\groundTruth,\classifier,\instanceSetGen,\impersonationGoalSet}(\adversary)$ of four attacks,  
using two new loss functions, \mdmaxloss and
\mdgrouploss, and the two baseline na\"{i}ve attacks, on the three image datasets.
We also ran attacks with \mdmaxloss along with the baselines on the \sst dataset.
\mahmood{we're evaluating
  \gbtrfull{} in this section, why redefine targeted and untargeted
  robustness again?} \weiran{addressed}
The performance of
attacks is computed on a per-image
basis: in experiments related to the new loss functions, 
$\experiment{\impRel}{\allowedPert,\groundTruth,\classifier,\instanceSetGen,\impersonationGoalSet}(\adversary)$ always uses a $\instanceSetGen$ that
samples $\instanceSet$ of size $\vert \instanceSet \vert=1$ uniformly.

\subsubsection{Results}
\label{sec:attack:loss:results}
\begin{figure}[t!]
\centerline{\includegraphics[width=0.95\columnwidth]{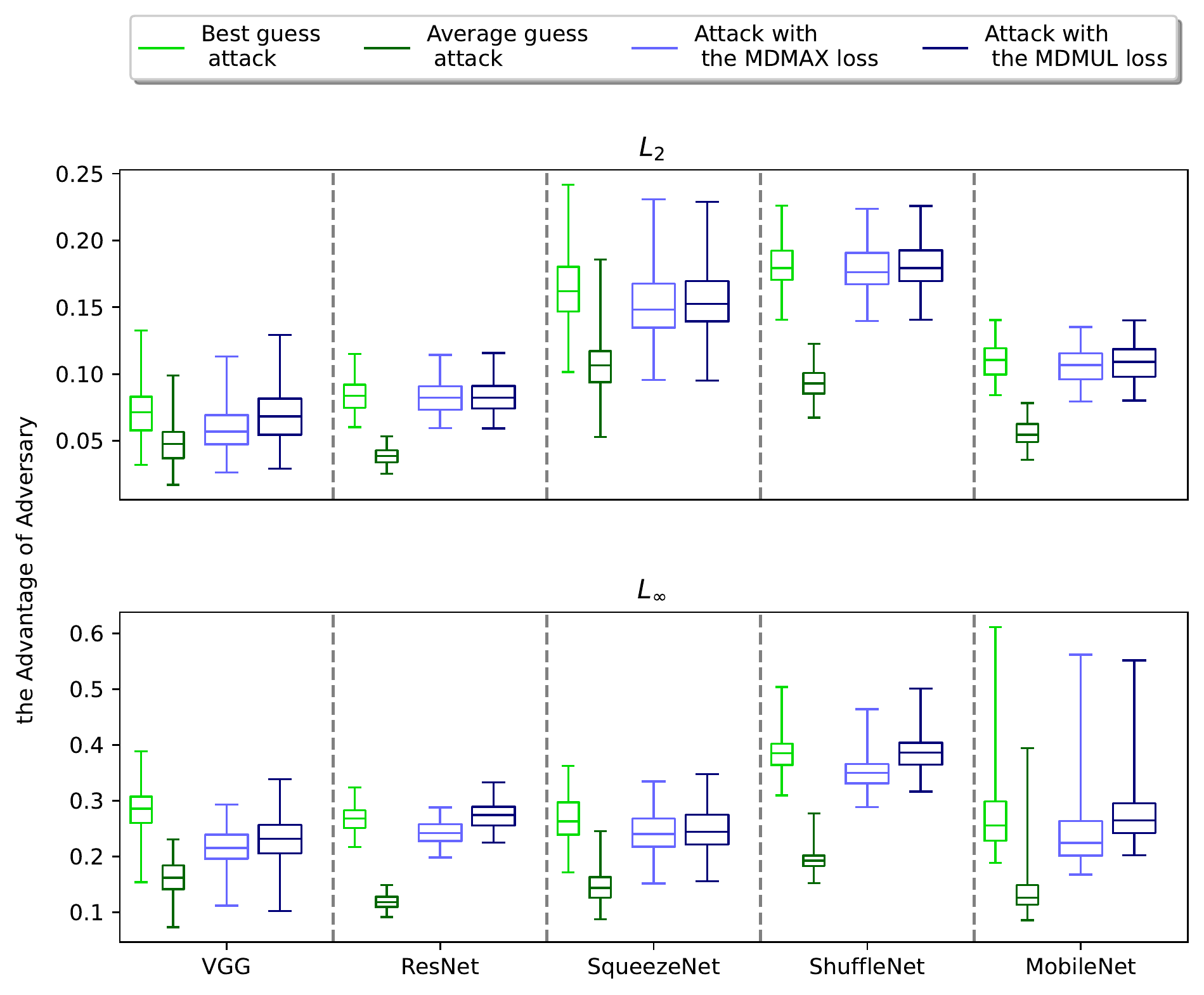}}
\caption{$\advantage{\impRel}{\allowedPert,\groundTruth,\classifier,\instanceSetGen,\impersonationGoalSet}(\adversary)$ of four attacks, two na\"{i}ve methods and
  two with new loss functions, on the \gtsrb{} dataset. Although the ranges of advantages of different attacks overlap,
 attacks with
  \mdmaxloss or \mdgrouploss, depicted as the wider boxes, usually have slightly lower $\advantage{\impRel}{\allowedPert,\groundTruth,\classifier,\instanceSetGen,\impersonationGoalSet}(\adversary)$
  than the best guess attack but always have higher advantages than the average guess attack.
  With each
  combination of $L_p$-norm and architecture, the performance of models varies only due to different randomly initialized weights, using seeds 0 -- 99.}
\label{fig:attacks:GTSRB}
\end{figure}

%
\begin{figure}[t!]
\centerline{\includegraphics[width=0.95\columnwidth]{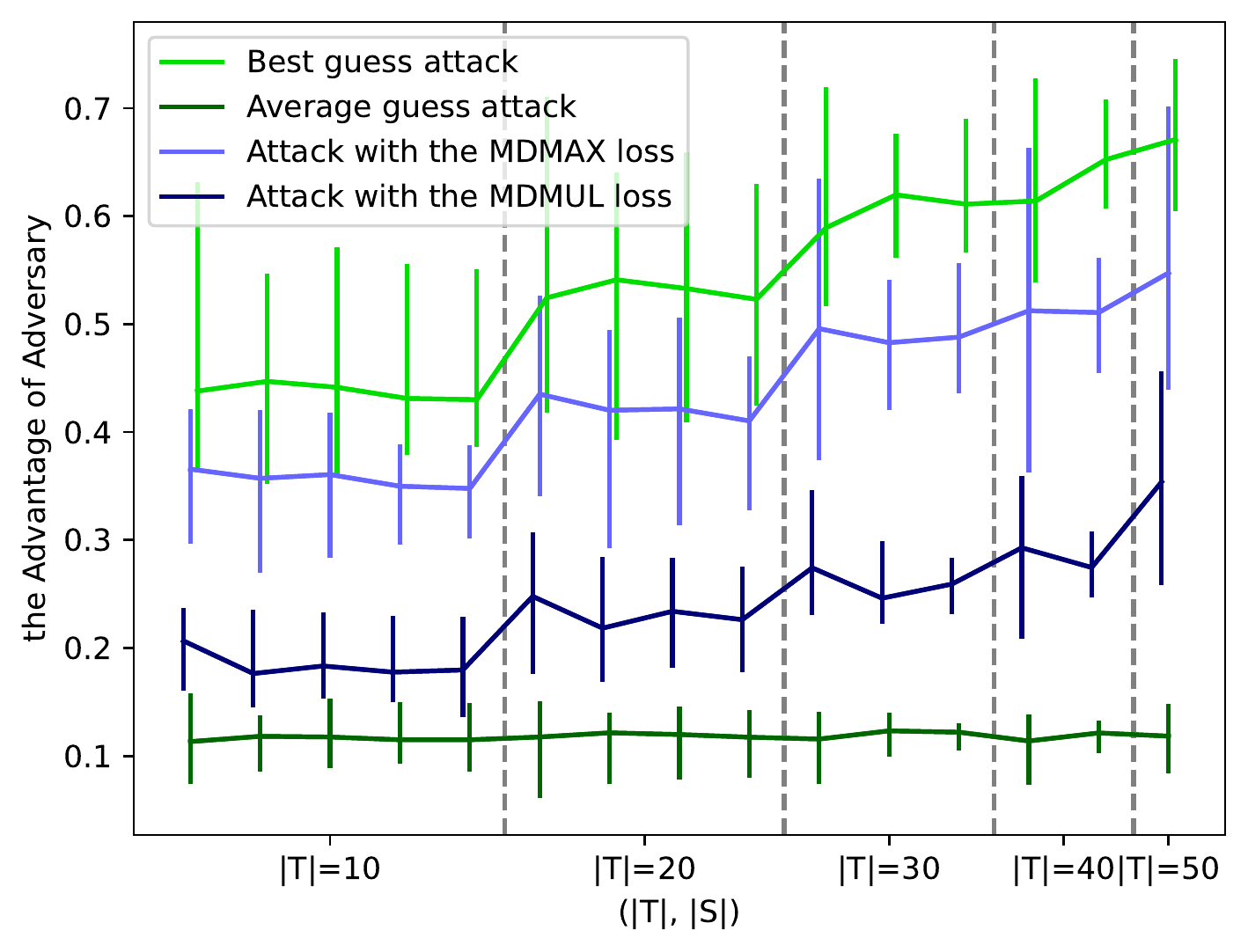}}
\caption{Advantages of four attacks, two na\"{i}ve methods
  and two with new loss functions, on the \pubfig{} dataset. Ranges
  are due to choices of $\sourceclassset$ and
  $\targetclassset$. Although the ranges of advantages of different attacks overlap,
  for each specific $(\sourceclassset,\targetclassset)$, attacks with
  \mdmaxloss or \mdgrouploss always have lower $\advantage{\impRel}{\allowedPert,\groundTruth,\classifier,\instanceSetGen,\impersonationGoalSet}(\adversary)$
  than the best guess attack but have higher advantages than the average
  guess attack.} 
\label{fig:attacks:2dface+}
\end{figure}

\begin{figure}[t!]
\centerline{\includegraphics[width=0.95\columnwidth]{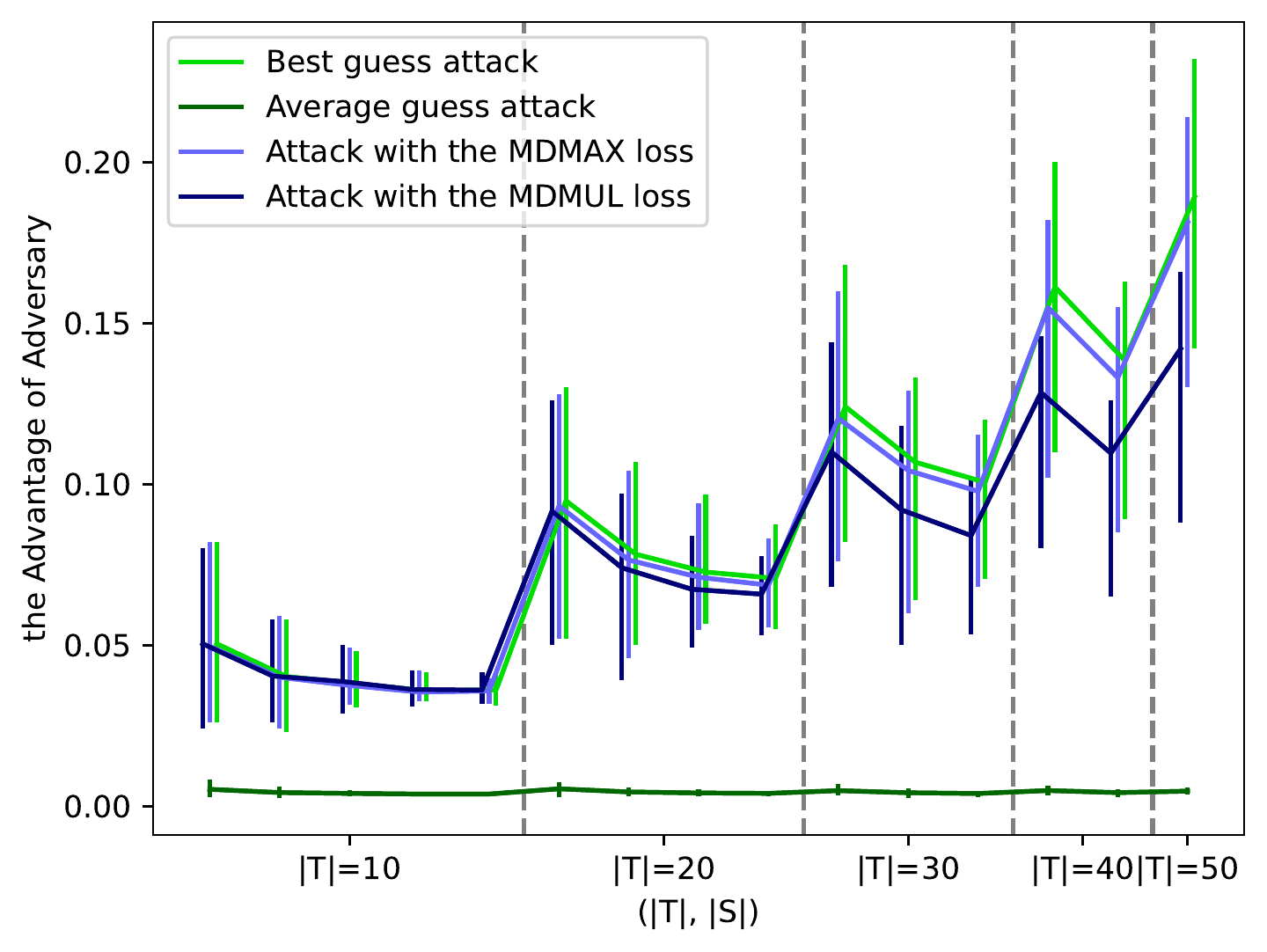}}
\caption{Advantages of four attacks, two na\"{i}ve methods
  and two with new loss functions, on the \imagenet dataset with $L_2$-norm. Ranges are due to choices of $\sourceclassset$ and
  $\targetclassset$. Although the ranges of advantages of different attacks overlap,
  for each specific $(\sourceclassset,\targetclassset)$, attacks with
  \mdmaxloss or \mdgrouploss usually have slightly lower $\advantage{\impRel}{\allowedPert,\groundTruth,\classifier,\instanceSetGen,\impersonationGoalSet}(\adversary)$
  than the best guess attack but always have higher advantages than the average
  guess attack.} 
\label{fig:attacks:2dL2+}
\end{figure}

\begin{figure}[t!]
\centerline{\includegraphics[width=0.95\columnwidth]{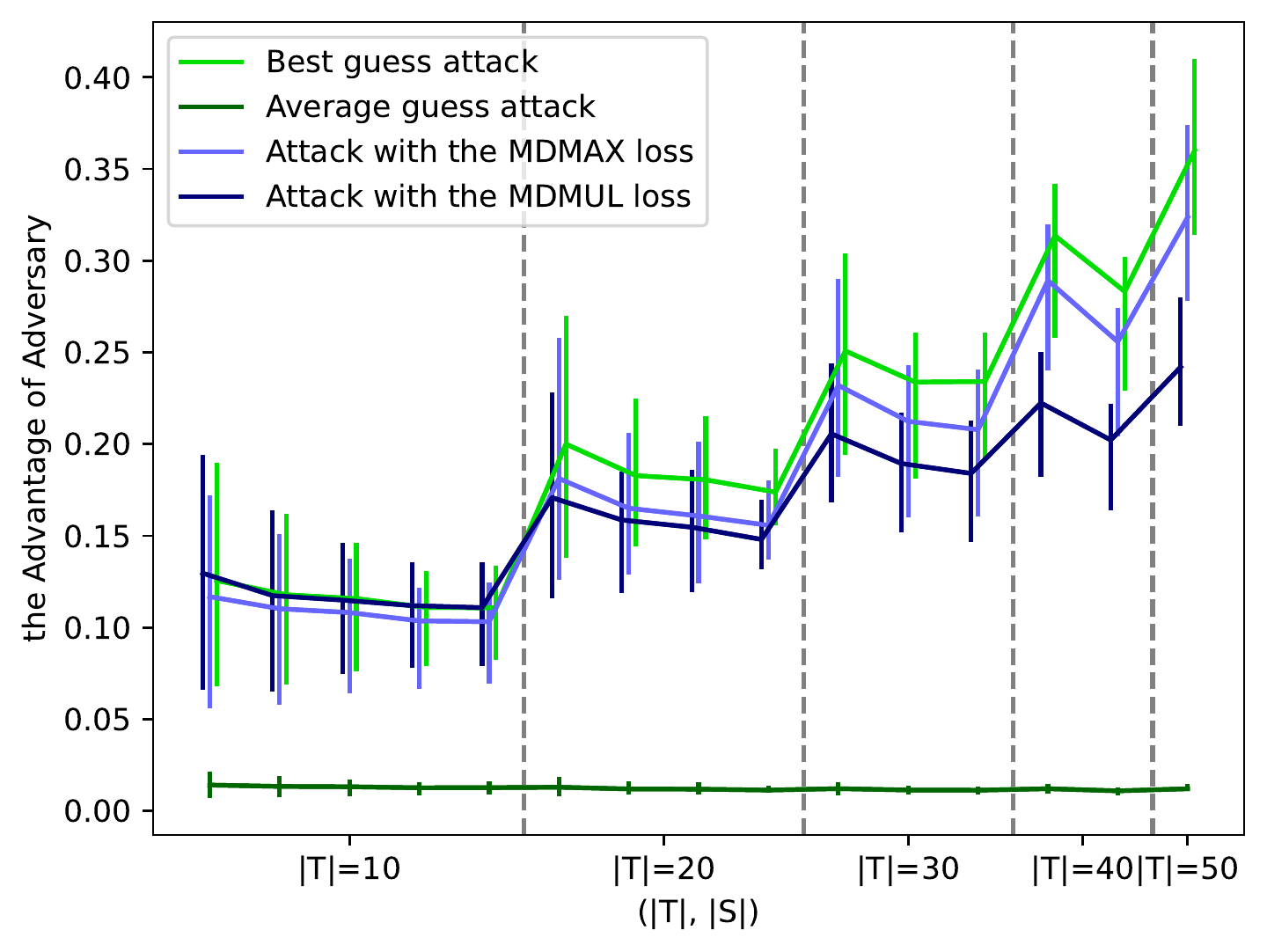}}
\caption{Advantages of four attacks, two na\"{i}ve methods
  and two with new loss functions, on the \imagenet{} dataset with
  $L_{\infty}$-norm. Ranges are due to choices of $\sourceclassset$ and
  $\targetclassset$. Although the ranges of advantages of different attacks overlap,
  for each specific $(\sourceclassset,\targetclassset)$, attacks with
  \mdmaxloss or \mdgrouploss usually have slightly lower $\advantage{\impRel}{\allowedPert,\groundTruth,\classifier,\instanceSetGen,\impersonationGoalSet}(\adversary)$
  than the best guess attack but always have higher advantages than the average
  guess attack.} 
\label{fig:attacks:2dLinf+}
\end{figure}

On the \gtsrb{} dataset, as illustrated in \figref{fig:signs}, $\vert
\targetclassset \vert$ ranges from 3 (for a 20 KPH sign) to 7 (for a
120 KPH sign). Compared with the best guess attacks, attacks with
\mdmaxloss or \mdgrouploss take intuitively and empirically one-third to one-seventh of
the time on \gtsrb{} (more details later).
As shown in \figref{fig:attacks:GTSRB}, 
the $\advantage{\impRel}{\allowedPert,\groundTruth,\classifier,\instanceSetGen,\impersonationGoalSet}(\adversary)$ of
attacks with new loss functions is $0.62\text{--}1.04\times$
that of
the best guess attacks.
Attacks with
 \mdmaxloss or \mdgrouploss take much less time than the best guess attacks, but obtain close
  $\advantage{\impRel}{\allowedPert,\groundTruth,\classifier,\instanceSetGen,\impersonationGoalSet}(\adversary)$. 
  The
average guess attacks take the same amount of time as attacks with
\mdmaxloss or \mdgrouploss on \gtsrb{}. 
The advantages of
attacks with the new loss functions are 
$1.04\text{--}2.56\times$ 
that of the
average guess attacks, i.e., always larger than the advantages of average guess attacks.
On the \pubfig{} and \imagenet datasets, $\vert \targetclassset \vert$
ranges from 10 to 50.  On the
\pubfig{} dataset, when $\vert \targetclassset \vert$ increases, 
 the $\advantage{\impRel}{\allowedPert,\groundTruth,\classifier,\instanceSetGen,\impersonationGoalSet}(\adversary)$
 of attacks with \mdmaxloss, attacks with \mdgrouploss,
and the best guess attacks also increase, whereas 
 the $\advantage{\impRel}{\allowedPert,\groundTruth,\classifier,\instanceSetGen,\impersonationGoalSet}(\adversary)$
 of the
average guess attacks stay about the
same. We observe the same phenomenon in these two datasets (shown in \figref{fig:attacks:2dface+}--\figref{fig:attacks:2dLinf+}).
Since $\vert \targetclassset \vert$ ranges from 10 to 50 on \pubfig{}
and \imagenet, the best guess attacks are 10 to 50 times slower than
attacks with \mdmaxloss or \mdgrouploss.
The $\advantage{\impRel}{\allowedPert,\groundTruth,\classifier,\instanceSetGen,\impersonationGoalSet}(\adversary)$ of new loss functions is $0.30\text{--}1.21\times$ as large as 
that of the best guess attacks on \pubfig{}, and $0.62\text{--}1.15\times$ on \imagenet.
The $\advantage{\impRel}{\allowedPert,\groundTruth,\classifier,\instanceSetGen,\impersonationGoalSet}(\adversary)$ of new loss functions is $1.19\text{--}6.22\times$ as large as 
that of the average guess attacks on \pubfig{}, and $7.37\text{--}41.53\times$ on \imagenet.
On the \sst dataset, attacks with \mdmaxloss obtain $\advantage{\impRel}{\allowedPert,\groundTruth,\classifier,\instanceSetGen,\impersonationGoalSet}(\adversary)$ that is not only larger than the average guess attacks but also unintuitively larger than the best guess attacks, as shown in \figref{fig:textattacks}. 
T-PGD uses a weighted sum of two parts as the loss function: the first part aims to induce misclassification, which we replace with \mdmaxloss, and the second part aims to increase the USE score. Using \mdmaxloss, T-PGD finds successful attacks earlier and thus spends more iterations to increase the USE score.
In summary, attacks with \mdmaxloss or \mdgrouploss always obtain larger
advantages
 than the average guess attacks do.


\begin{figure}[t!]
\centerline{\includegraphics[width=0.95\columnwidth]{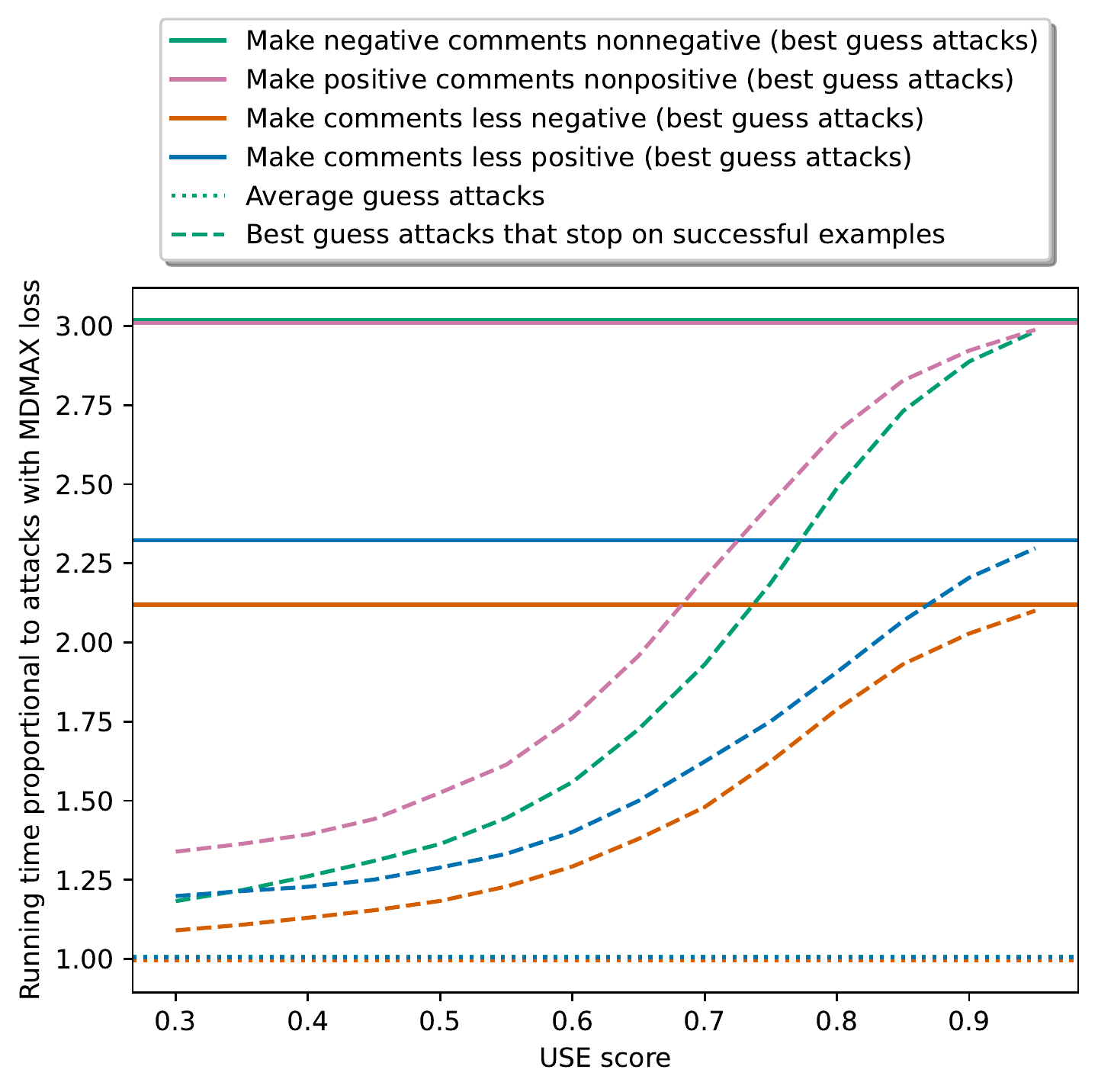}}
\caption{Running time of two baselines and attacks with \mdmaxloss on \sst. Attacks with \mdmaxloss take about the same time as the average guess attacks. As the USE score increases, finding successful attacks becomes harder, and best guess attacks take much more time than attacks with \mdmaxloss.  }
\label{fig:texttime}
\end{figure}

\begin{figure}[t!]
\small
\caption{Average time (seconds) to perturb one batch of images. Intuitively attacks with $\elmdmax$ 
or $\elmdgroup$ are faster by a factor of $\vert
\targetclassset \vert$ than the best guess attacks. We measured the average running time on a RTX 3090 GPU, with the largest $\vert
\targetclassset \vert$ for each dataset: 50 for \imagenet and
\pubfig{}, and 7 for \gtsrb{}.}
\begin{center}
\begin{tabular}{|@{\hskip 0.01in}c@{\hskip 0.01in}|@{\hskip 0.01in}c@{\hskip 0.1in}c@{\hskip 0.01in}|@{\hskip 0.05in}r@{\hskip 0.05in}r@{\hskip 0.05in}r@{\hskip 0.05in}r@{\hskip 0.05in}|}
\hline

\textit{Dataset}&$\mathit{L_p}$&\textit{architecture}&\textit{Best}&\textit{Average}&\textit{Attack}& $\textit{Attack}$\\
& &&\textit{ guess}&\textit{ guess}&\textit{with}& \textit{with}\\
& \textit{norm}&&\textit{attack}&\textit{attack}&$\mathit{\elmdmax}$& $\mathit{\elmdgroup}$\\
\hline

\imagenet&$L_2$&--&279.44&5.59&5.07&5.67\\
\cline{2-7}
&$L_{\infty}$&--&280.89&5.61&5.06&5.51\\
\hline
\pubfig{}&--&--&845.93&16.92&15.33&16.48\\
\hline
&&VGG&18.54&2.65&2.48&2.59\\
&&ResNet&18.01&2.57&2.39&2.51\\
&$L_2$&SqueezeNet&11.27&1.61&1.57&1.59\\
&&ShuffleNet&17.57&2.51&2.37&2.51\\
\gtsrb{}&&MobileNet&19.88&2.84&2.44&2.53\\
\cline{2-7}
&&VGG&17.18&2.45&2.28&2.32\\
&&ResNet&17.04&2.43&2.24&2.35\\
&$L_{\infty}$&SqueezeNet&11.16&1.58&1.47&1.64\\
&&ShuffleNet&16.84&2.41&2.25&2.48\\
&&MobileNet&18.06&2.58&2.37&2.57\\
\hline

\end{tabular}
\end{center}
\label{tab:time}
\end{figure}
 
To empirically verify that the two new loss functions reduce the
computation time of attacks, we measured the average running time to
perturb one batch of images on an RTX 3090 GPU. We used 10 images from
\imagenet, 64 images from \pubfig{}, or 512 images from  
\gtsrb{}, as one batch of images. We used the largest $\vert
\targetclassset \vert$ for each dataset: 50 for \imagenet and \pubfig{}, and 7 for \gtsrb{}. 
We compared four attacks: the best guess attacks, the average guess attacks, attacks with \mdgrouploss, and attacks with \mdmaxloss.
The detailed results are shown in \figref{tab:time}.
The results confirmed our hypothesis: attacks with \mdgrouploss or
\mdmaxloss are faster than the best guess attacks by a factor of
$\vert 
\targetclassset \vert$.
We also noticed that if adversaries choose to perturb images by
instances, instead of conventional batches, they might save time to
run the best guess attacks by stopping perturbing an image that they
have already succeeded in perturbing. While empirically such a variant of
the best guess attacks save time for the adversaries, the time needed
by this variant is still much larger than attacks with \mdgrouploss or
\mdmaxloss.
The detailed results are shown in \figref{tab:time}.
We observe the similar results as shown in \figref{fig:texttime}:
attacks with \mdmaxloss take about the same amount of time as average guess attacks, and they take much less time especially when the USE score is high.

\begin{figure}[t!]
\small
\caption{Average time (seconds) to perturb one image. Intuitively attacks with $\elmdmax$ 
or $\elmdgroup$ are faster by a factor of $\vert
\targetclassset \vert$ than the best guess attacks. We measured the average running time on a RTX 3090 GPU, with the largest $\vert
\targetclassset \vert$ for each dataset: 50 for \imagenet and \pubfig{}, and 7 for \gtsrb{}. }
\begin{center}
\resizebox{\columnwidth}{!}{
\begin{tabular}{|@{\hskip 0.01in}c@{\hskip 0.01in}|@{\hskip 0.01in}c@{\hskip 0.1in}c@{\hskip 0.01in}|@{\hskip 0.05in}r@{\hskip 0.05in}r@{\hskip 0.05in}r@{\hskip 0.05in}r@{\hskip 0.05in}|}
\hline

\textit{Dataset}&$\mathit{L_p}$&\textit{architecture}&\textit{Best guess}&\textit{Best guess}&\textit{Attack}& $\textit{Attack}$\\
& &&\textit{ attack}&\textit{attack }&\textit{with}& \textit{with}\\
& \textit{norm}&&\textit{}&\textit{(until success)}&$\mathit{\elmdmax}$& $\mathit{\elmdgroup}$\\
\hline
\imagenet&$L_2$&--&205.04&177.40&4.09&4.19\\
\cline{2-7}
&$L_{\infty}$&--&189.06&148.15&3.87&3.92\\
\hline
\pubfig{}&--&--&135.96&58.21&2.58&7.77\\
\hline
&&VGG&8.73&8.19&1.27&1.49\\
&&ResNet&10.02&9.28&1.44&1.64\\
&$L_2$&SqueezeNet&10.06&7.47&1.49&1.54\\
&&ShuffleNet&16.54&13.70&2.30&2.45\\
\gtsrb{}&&MobileNet&15.55&13.90&2.25&2.43\\
\cline{2-7}
&&VGG&8.18&5.94&1.24&1.54\\
&&ResNet&9.92&7.35&1.41&1.61\\
&$L_{\infty}$&SqueezeNet&10.61&8.29&1.45&1.61\\
&&ShuffleNet&15.78&13.52&2.16&2.31\\
&&MobileNet&14.97&13.75&2.27&2.40\\
\hline
\end{tabular}}
\end{center}
\label{tab:smarttime}
\end{figure}

\begin{myblock}{Takeaways (Loss Functions)}
Attacks with \mdmaxloss or \mdgrouploss
achieve comparable or slightly lower 
$\advantage{\impRel}{\allowedPert,\groundTruth,\classifier,\instanceSetGen,\impersonationGoalSet}(\adversary)$
than the best guess attacks, consume markedly less time and are
markedly more efficient, finding more attacks per time unit. Attacks with \mdmaxloss or \mdgrouploss
 consume the same amount of time as the average guess attacks and have much higher 
 $\advantage{\impRel}{\allowedPert,\groundTruth,\classifier,\instanceSetGen,\impersonationGoalSet}(\adversary)$. 
The MDMUL loss
and \mdmaxloss boost the efficiency of attacks in the attack
scenarios we tried.
\end{myblock}

\subsection {Attack Strategies}
\label{sec:attacks:strategy}

The previous section (\secref{sec:attacks:loss}) introduced loss functions that
increased the efficiency of evasion attacks
that perturb a single input toward a set containing multiple
target classes. In contrast, this section introduces
strategies that can be used to increase efficiency when
there are also multiple inputs to be perturbed.

Previous works evaluate attacks by either the success rate
$\advantage{\impRel}{\allowedPert,\groundTruth,\classifier,\instanceSetGen,\impersonationGoalSet}(\adversary)$
(e.g., \cite{ICML22CGD,icml20:autopgd,iclr18:PGD,iclr15:fgsm,iclr17:i-fgsm,iclr14:lbfgs,cvpr16:deepfool,cvpr18:momentum,Euro16:JSMA,ICMLA17:semantic,nips19:sparse,ICML18:spsa,ccs16:eyeglasses})
or $L_{p}$-norm distance \allowedPert
(e.g.,~\cite{Oakland17:CarliniWagner,iclr16:featureadv,Oakland20:HopSkip,ICML18:noise}).
These metrics measure the per-input-sample performance of
attacks. 
That is, $\vert \instanceSet \vert=1$ for $\instanceSetGen$.
However, as described in \secref{sec:metric}, adversaries
could face scenarios where they need to cause 
each of several images that belong to different classes to be
incorrectly classified into several other different classes. 
In scenarios such as
the burglary example, 
more than one burglar may impersonate the bank staff, and more than one staff member needs to be impersonated
since no staff member can grant access individually.
Attacks \adversary need to maximize
$\advantage{\impRel}{\allowedPert,\groundTruth,\classifier,\instanceSetGen,\impersonationGoalSet}(\adversary)$
when $\vert \instanceSet \vert>1$ and each
$\impersonationGoal{\impersonationGoalIdx} \in \impersonationGoalSet$
is a surjective function mapping classes \sourceclassset to a target
set \targetclassset of classes where $\sourceclassset \cap
\targetclassset = \emptyset$.  
  A major
obstacle to computing $\advantage{\impRel}{\allowedPert,\groundTruth,\classifier,\instanceSetGen,\impersonationGoalSet}(\adversary)$ when $\vert \instanceSet \vert>1$ 
is that whether an attack can successfully perturb an input instance $\instance \in \instanceSet$ as a target class $\target \in \targetclassset$
remains unknown until
an exhaustive attack is fully performed. 
Suppose $\instance$ is a specific instance in $\instanceSet$ generated by $\instanceSetGen$, $\target$ is a specific target class $\target \in \targetclassset$,
$\sourceclass=\groundTruth(\instance)$, $\tilde{\impersonationGoalSet}=\left\{\{(\sourceclass,
\targetclass{})\}\right\}$ and $\tilde{\instanceSetGen}$ 
is an algorithm producing instance
 $\tilde{\instanceSet}=\{\instance\}$ 
($\vert \instanceSet \vert>1$ for $\instanceSetGen$ and $\vert \tilde{\instanceSet} \vert=1$ for $\tilde{\instanceSetGen}$).
If the adversary $\adversary$ can estimate the pairwise success rate 
$\advantage{\impRel}{\allowedPert,\groundTruth,\classifier,\tilde{\instanceSetGen},\tilde{\impersonationGoalSet}}(\tilde{\adversary})$ to perturb \instance
to \target,
it may choose the $\instance$ and $\target$ pairs accordingly to focus its attack, 
obtaining a higher probability to have $\experiment{\impRel}{\allowedPert,\groundTruth,\classifier,\instanceSetGen,\impersonationGoalSet}(\adversary)$
return 1
and thus maximizing 
$\advantage{\impRel}{\allowedPert,\groundTruth,\classifier,\instanceSetGen,\impersonationGoalSet}(\adversary)$ by definition.
We will now introduce three
  strategies to
estimate $\advantage{\impRel}{\allowedPert,\groundTruth,\classifier,\tilde{\instanceSetGen},\tilde{\impersonationGoalSet}}(\tilde{\adversary})$,
the pairwise success rate 
to perturb each $\instance \in \instanceSet$ as each $\target \in \targetclassset$.

\paragraph{Estimate by Computing a Prior from Validation Set}
\adversary can launch targeted attacks using a validation set,
perturbing input instances associated with each $\sourceclass \in \sourceclassset$ as each 
$\target \in \targetclassset$.
In the bank burglary example, the burglars might collect images of staff ahead of time to construct the validation set.
\adversary can
 compute a prior probability of perturbing input instances associated with each $\sourceclass \in \sourceclassset$ as each 
$\target \in \targetclassset$.
More formally, $\tilde{\adversary}$ is trying to transform a random instance of \sourceclass to \target, as specified by $\tilde{\impersonationGoalSet}$. 
 \adversary can compute 
$\advantage{\impRel}{\allowedPert,\groundTruth,\classifier,\instanceSetGen,\tilde{\impersonationGoalSet}}(\tilde{\adversary})$,
and use it as an estimate of
 $\advantage{\impRel}{\allowedPert,\groundTruth,\classifier,\tilde{\instanceSetGen},\tilde{\impersonationGoalSet}}(\tilde{\adversary})$.
This approach does not require $\instance$, the actual instance to be perturbed, to estimate.
In the burglary example, using this approach, the burglars \adversary do not need to maintain the same poses before the facial recognition camera
when they try to impersonate different staff members.
  When \adversary executes,
\adversary always tries the $(\sourceclass,\target)$ class pair with highest prior 
$\advantage{\impRel}{\allowedPert,\groundTruth,\classifier,\instanceSetGen,\tilde{\impersonationGoalSet}}(\tilde{\adversary})$
first. It iterates over input instances associated with class $\sourceclass$ and performs targeted
attacks towards class $\target$ until a successful perturbation has been
found or all images have been tried. Then it repeats the above
process by picking the $(\sourceclass,\target)$ class pair with the next highest prior,
and skipping such a pair if a successful perturbation targeting $\target$ has
already been found.

\paragraph{Estimate by \md Loss Without Perturbation}
As discussed in \secref{sec:attacks:loss}, the smaller the
\md loss, the closer the attack is to succeeding. \adversary can
perform one forward propagation for every input instance $\instance \in \instanceSet$
  to be perturbed to
get the logits.  
Then \adversary can compute the \md loss towards each
target class $\target \in \targetclassset$. 
The smaller the \md loss is, the larger the estimated $\advantage{\impRel}{\allowedPert,\groundTruth,\classifier,\tilde{\instanceSetGen},\tilde{\impersonationGoalSet}}(\tilde{\adversary})$ is.
When \adversary is carried out, \adversary repeatedly selects $(\instance,\target)$ pairs
with the next smallest \md loss, and it skips pairs where a successful
perturbation has been found targeting class $\target$.

\paragraph{Estimate by \md Loss After One Attack Iteration}
For every input instance $\instance \in \instanceSet$ and each target class $\target \in \targetclassset$, \adversary  perform one
iteration of the attack $\tilde{\adversary}$~\cite{iclr15:fgsm,iclr17:i-fgsm} before computing
the \md loss. 
The smaller the \md loss is, the larger the estimated $\advantage{\impRel}{\allowedPert,\groundTruth,\classifier,\tilde{\instanceSetGen},\tilde{\impersonationGoalSet}}(\tilde{\adversary})$ is.
When \adversary is carried out,
 \adversary repetitively selects $(\instance,\target)$ pairs with the
next smallest \md loss, and it skips pairs where a successful perturbation
has been found targeting class $\target$. 

\medskip

\adversary do not have to use the above three strategies
independently. Although the \md loss is not a probability estimate, we expect it to be
inversely correlated with the probability estimate of pairwise attack
success $\advantage{\impRel}{\allowedPert,\groundTruth,\classifier,\tilde{\instanceSetGen},\tilde{\impersonationGoalSet}}(\tilde{\adversary})$. 
Given an input instance $\instance \in \instanceSet$
from class $\groundTruth(\instance) = \sourceclass \in \sourceclassset$, and $\target \in \targetclassset$, \adversary
can compute the prior, $(\sourceclass,\target)$, $\advantage{\impRel}{\allowedPert,\groundTruth,\classifier,\instanceSetGen,\tilde{\impersonationGoalSet}}(\tilde{\adversary})$,
and \md loss of $(\instance,\target)$, with or
without adding perturbations, as above-mentioned. One
approach to combine these two values is to construct a product as 
$(1-prior)*\ell_{\mathit{\md}}$ for each $(\instance,\target)$ pair. 
The smaller the product, the larger the estimated $\advantage{\impRel}{\allowedPert,\groundTruth,\classifier,\tilde{\instanceSetGen},\tilde{\impersonationGoalSet}}(\tilde{\adversary})$ is.
When \adversary is carried out,
\adversary repeatedly
selects $(\instance,\target)$ pairs with the next smallest product and skips pairs
where a successful perturbation has been found targeting class
$\target$. Other approaches to combine strategies might also work.

\subsubsection{Experiment Setup}
To illustrate the new loss strategies' improvements of
$\advantage{\impRel}{\allowedPert,\groundTruth,\classifier,\instanceSetGen,\impersonationGoalSet}(\adversary)$, we used the same threat model, datasets and benchmarks as we did in \secref{sec:metric:setup}. In addition, we used the baselines and measurement process below.

\subsubsection*{Baselines}
We created a baseline strategy to compare with attack strategies proposed in \secref{sec:attacks:strategy}.
As we introduced in \secref{sec:metric},
when $\vert \instanceSet \vert>1$,
\adversary may consider goals
for which, e.g., each
$\impersonationGoal{\impersonationGoalIdx} \in \impersonationGoalSet$
is a surjective function mapping classes $\sourceclassset$ to a target
set $\targetclassset$ of classes where $\sourceclassset \cap
\targetclassset = \emptyset$.
For all $\instance \in \instanceSet$ and $\target \in \targetclassset$,
the baseline strategy randomly selects $(\instance,\target)$ pairs without replacement
and skips $(\instance,\target)$ pairs where a successful misclassification
targeting a $\target$ has been made.

\subsubsection*{Measurement Process}
\label{sec:attacks:strategies:process}

 We evaluated the attack strategies 
 on \pubfig{} and \imagenet using the bank burglary scenario.
 We implemented \instanceSetGen 
  for each test $\impersonationGoalSet$ on different datasets. 
  However, different from \secref{sec:metric:setup} and \secref{sec:attacks:loss:process},
 we used $\vert \instanceSet \vert>1$ as we introduced in \secref{sec:metric:space},
and
$\impersonationGoal{\impersonationGoalIdx} \in \impersonationGoalSet$
is a surjective function. 
 The size of the codomain of the surjective function is $K$, $K>1$.
 We compared the baseline strategy with five strategies: three strategies listed in \secref{sec:attacks:strategy}, and two strategies that are combinations of the previous three (also mentioned in \secref{sec:attacks:strategy}).
 
The strategy that estimates pairwise success rates of attacks,
$\advantage{\impRel}{\allowedPert,\groundTruth,\classifier,\tilde{\instanceSetGen},\tilde{\impersonationGoalSet}}(\tilde{\adversary})$,
 by computing a prior, $\advantage{\impRel}{\allowedPert,\groundTruth,\classifier,\instanceSetGen,\tilde{\impersonationGoalSet}}(\tilde{\adversary})$
,
 requires a validation set besides the data samples we used for evaluation.
\pubfig{} has its own validation set.
However, \imagenet only has a
testing set, so we randomly split it by a 2:1 ratio, as
\pubfig{} has.

We sampled images from the testing sets to evaluate the attack strategies.
For each of the ($\sourceclassset$, $\targetclassset$) pairs
described above, we first randomly selected one $\instance \in \instanceSet$ from each $\sourceclass \in \sourceclassset$, for a total of $\vert \sourceclassset \vert$ images. Then
we tried to perturb these images as any $K$
diverse classes. 
  In the bank burglary scenario,  $K$ is the least number of staff that need to agree to grant access to the treasury.

According to the estimated pairwise success rates,
$\advantage{\impRel}{\allowedPert,\groundTruth,\classifier,\tilde{\instanceSetGen},\tilde{\impersonationGoalSet}}(\tilde{\adversary})$,
we select the next most likely vulnerable 
$(\instance,\target)$ pairs to launch targeted attacks. 
We launch \autopgd with \md loss
for each $(\instance,\target)$ pair and counted it as one attempt.  
The less attempts needed to find $K$ diverse misclassifications $\in \targetclassset$,
the more often $\experiment{\impRel}{\allowedPert,\groundTruth,\classifier,\instanceSetGen,\impersonationGoalSet}(\adversary)$ returns 1,
and the larger $\advantage{\impRel}{\allowedPert,\groundTruth,\classifier,\instanceSetGen,\impersonationGoalSet}(\adversary)$ is.
As defined in \secref{sec:metric:definition},
$\advantage{\impRel}{\allowedPert,\groundTruth,\classifier,\instanceSetGen,\impersonationGoalSet}(\adversary)$ is the probability that 
$\experiment{\impRel}{\allowedPert,\groundTruth,\classifier,\instanceSetGen,\impersonationGoalSet}(\adversary)$ returns 1.

Assuming at least $K$ diverse misclassifications could always be found, we
compared attack strategies by the number of attempts needed to find $K$
diverse misclassifications. We ensured the assumption always holds
when selecting 1000 sets of $\vert \sourceclassset \vert$ images for
each ($\sourceclassset$, $\targetclassset$) pair. We used $K \in
\{5,10,15,20,25,30,35,40,45\}$ on \pubfig{} and $K \in
\{2,3,4,5,6,7,8\}$ on \imagenet. We eliminated choices of $K$ if
$K>\vert \targetclassset \vert$ or the probability that the
assumptions holds is smaller than $1\%$. For choices of $K$ that are
larger than the ones we listed above, the probability that the
assumptions hold was always smaller than $1\%$. 

\subsubsection{Results}
\label{sec:attacks:strategies:results}
\begin{figure}[t!]
\centerline{\includegraphics[width=0.95\columnwidth]{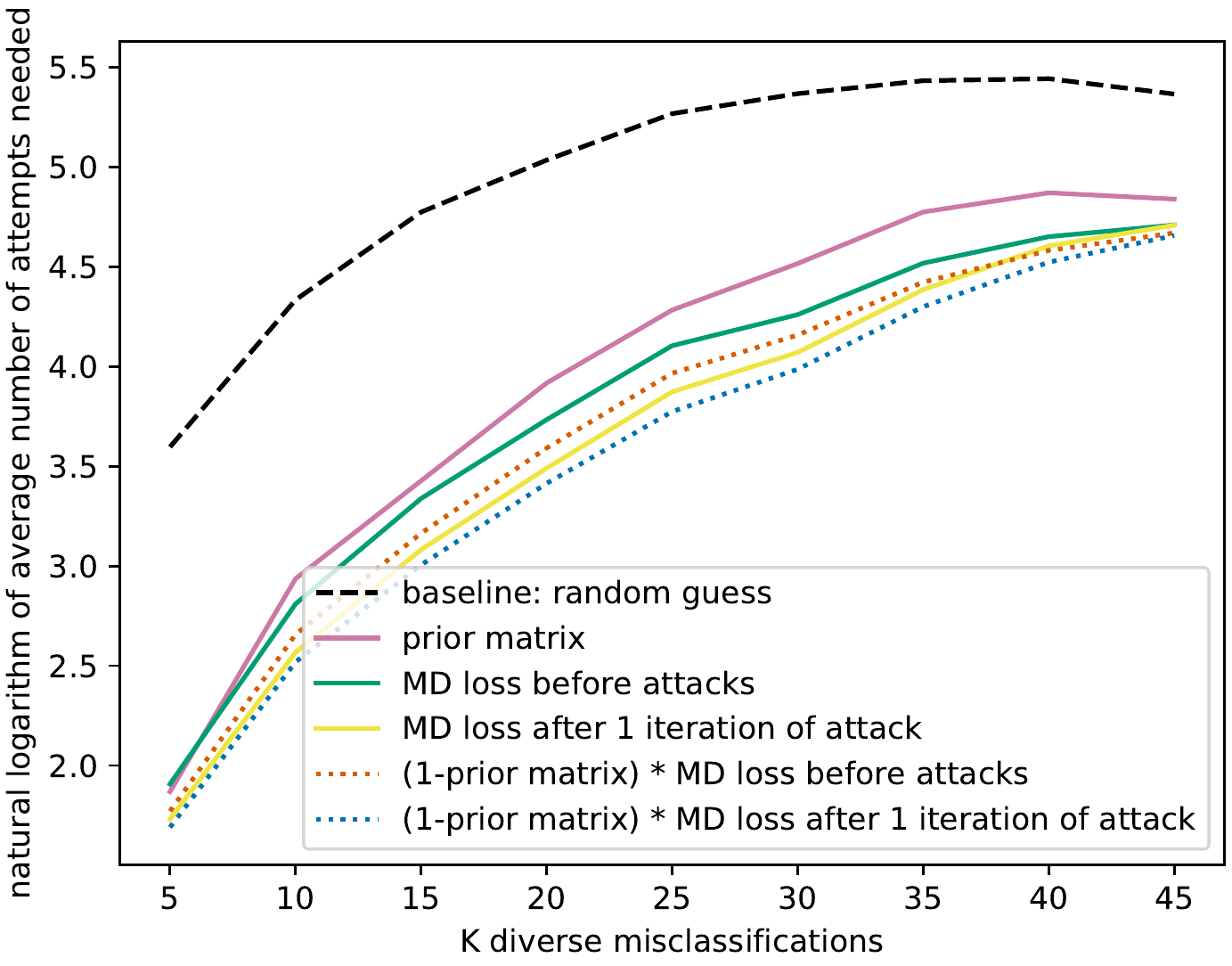}}
\caption{The natural logarithm of the average number of attempts needed by
  attack strategies to find $K$ diverse misclassifications,
  using eyeglasses attacks on the \pubfig{} dataset. Average is taken
  over images and choices of $(\sourceclassset,\targetclassset)$.}
\label{fig:strategies:face}
\end{figure}

\begin{figure}[t!]
\centerline{\includegraphics[width=0.95\columnwidth]{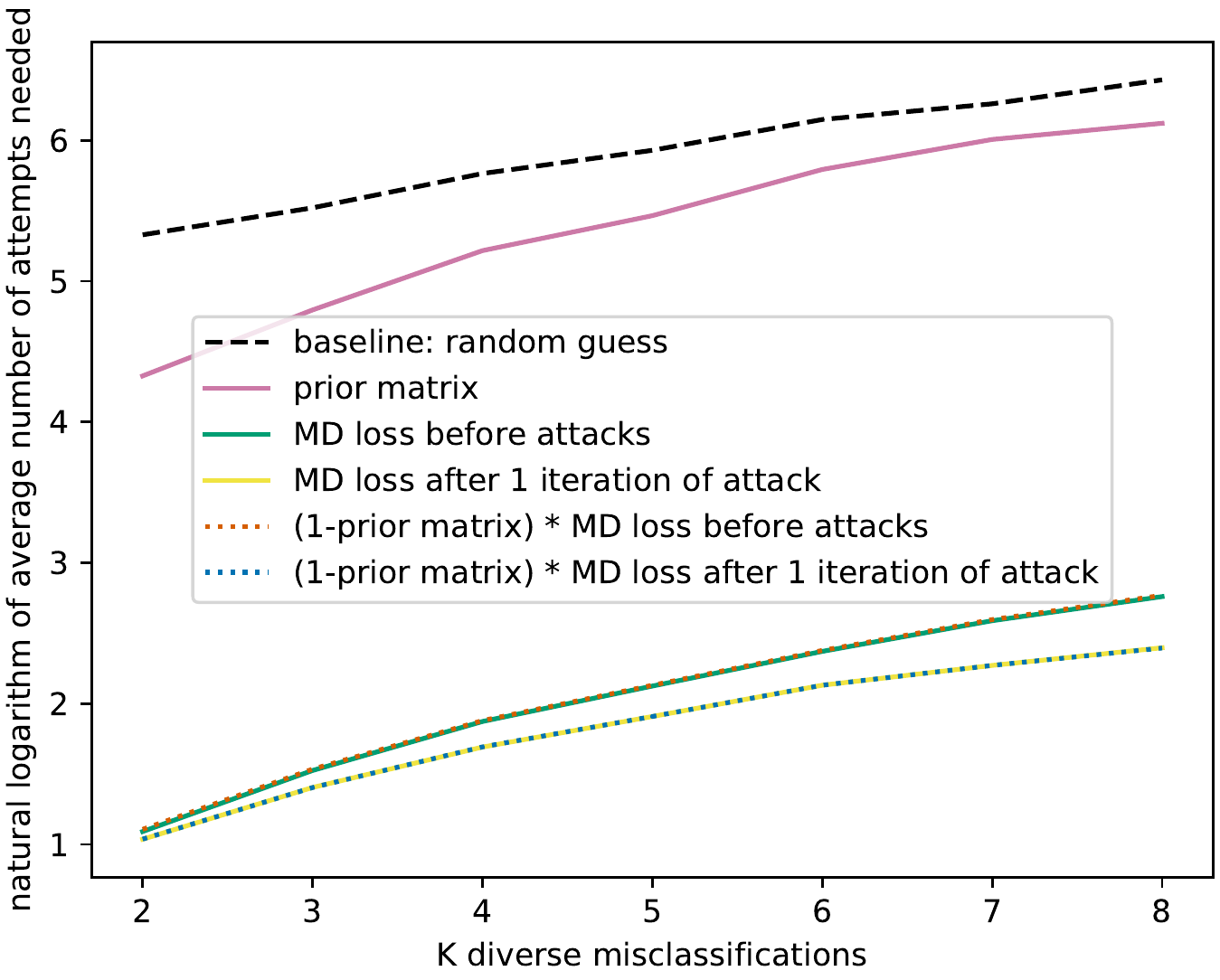}}
\caption{The natural logarithm of the average number of attempts needed by
  attack strategies to find K diverse misclassifications,
  using $L_2$ attacks on the \imagenet dataset. Average is taken over
  images and choices of $(\sourceclassset,\targetclassset)$.}
\label{fig:strategies:l2}
\end{figure}

\begin{figure}[t!]
\centerline{\includegraphics[width=0.95\columnwidth]{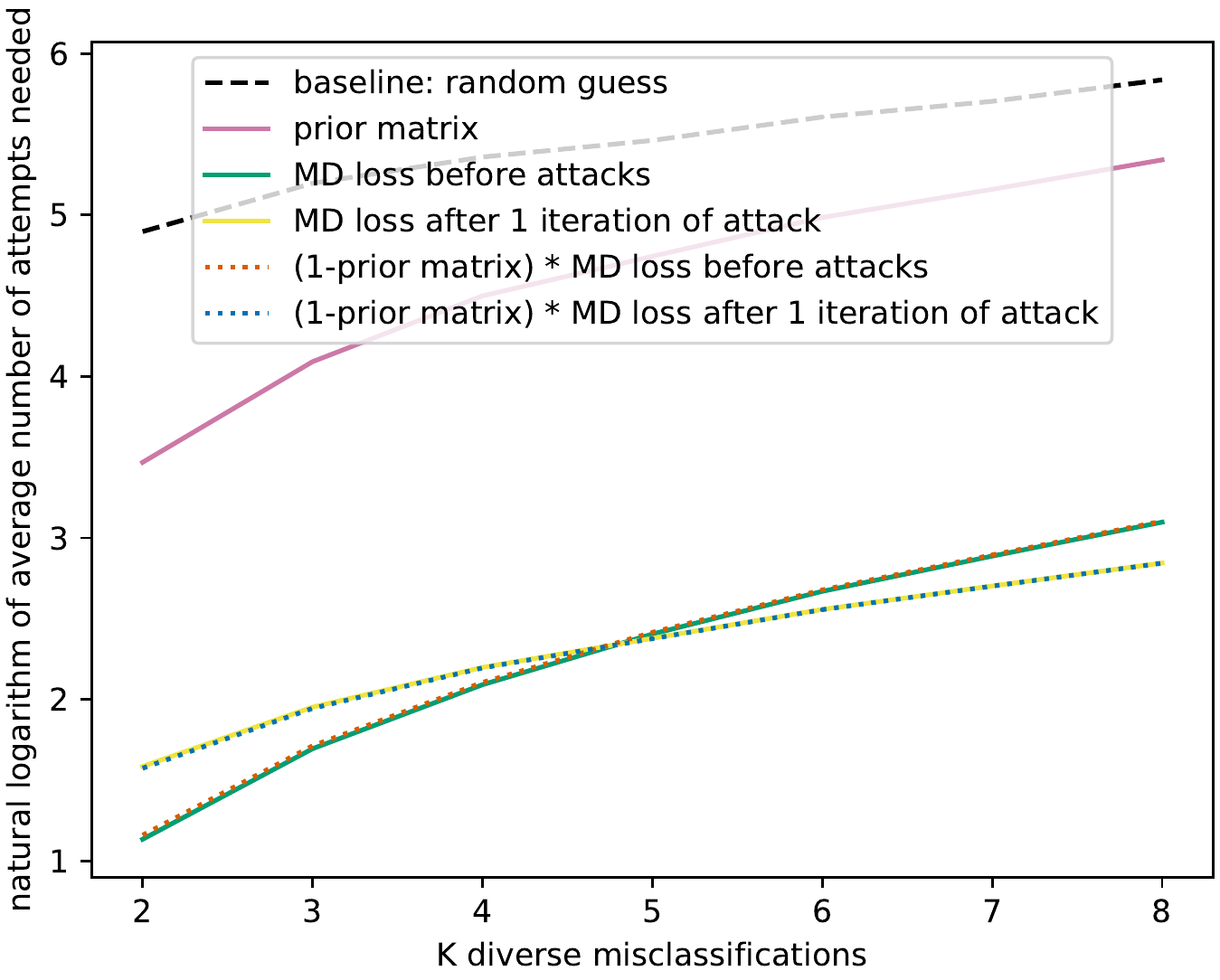}}
\caption{The natural logarithm of the average number of attempts needed by
  attack strategies to find $K$ diverse misclassifications,
  using $L_{\infty}$ attacks on the \imagenet dataset. Average is
  taken over images and choices of
  $(\sourceclassset,\targetclassset)$.}
\label{fig:strategies:linf}
\end{figure}

Now we turn to empirically demonstrate that the new attack strategies 
help to increase $\advantage{\impRel}{\allowedPert,\groundTruth,\classifier,\instanceSetGen,\impersonationGoalSet}(\adversary)$,
 where $\vert \instanceSet \vert>1$ and  $\impersonationGoal{\impersonationGoalIdx} \in \impersonationGoalSet$
is a surjective function.
The detailed results are shown in \figrefs{fig:strategies:face}{fig:strategies:linf}. 
 On all
datasets and benchmarks, estimating the success rate only by the prior
matrix is more efficient than random guess but less efficient than all
other methods. This strategy needs only 13.48--60.25\% of the number
of attempts needed by the baseline on \pubfig{}, 23.84--92.63\% on
\imagenet with $L_2$-norm, and 10.92--89.00\% on \imagenet with
$L_{\infty}$-norm.
This strategy has larger $\advantage{\impRel}{\allowedPert,\groundTruth,\classifier,\instanceSetGen,\impersonationGoalSet}(\adversary)$ than random guess.

Estimating the pairwise success rate $\advantage{\impRel}{\allowedPert,\groundTruth,\classifier,\tilde{\instanceSetGen},\tilde{\impersonationGoalSet}}(\tilde{\adversary})$ 
with MD loss after 1 iteration of the attack
is less efficient than without any perturbation only when using
small values of K and $L_{\infty}$ norm on the \imagenet
dataset. Combining the prior with other strategies on the
\pubfig{} dataset creates more efficient strategies. We verified that
these conclusions have statistical significance, using the same Wilcoxon
signed-rank test as Lin et al.\ used to compare success rates of
attacks~\cite{ICML22CGD}. Combining the prior with other
strategies on the \imagenet dataset has minimal effects on the
efficiency.
The most efficient strategy needs only 12.41--49.19\% of the
number of attempts needed by the baseline on \pubfig{}, 0.79--4.53\% on
\imagenet with $L_2$-norm, and 1.26--9.40\% on \imagenet with
$L_{\infty}$-norm. 

We additionally evaluated the strategies on
\imagenet with two variants of the current setup, each with an additional constraint on 
$\adversary$. In the first variant, 
we randomly selected a five-class subset of $\sourceclassset$ as managers, allowing the vault to be opened only if 1) at least $K$ staff members agree and 2) at least one of these $K$ staff members is a manager.
We modified $\adversary$ so it would still try to impersonate any of the staff members until $K-1$ success, and then only try to impersonate any of the managers if no manager has been impersonated.
The relationship between attack strategies remains the same: previously more efficient strategies are still more efficient in this setting. The most efficient strategy needs only 0.89--3.41\% of the
number of attempts needed by the baseline with $L_2$-norm, and 2.81--9.40\% with
$L_{\infty}$-norm. 

In the second variant, we required that the face-recognition system recognizes all $K$ staff members simultaneously and hence each burglar might impersonate once at most. This is in contrast with the previous two settings, where the face-recognition system takes one face at a time, and the same burglar may impersonate more than one staff members. 
The relationship between attack strategies still remains the same. 
The most efficient strategy needs only 0.95--3.55\% of the
number of attempts needed by the baseline with $L_2$-norm, and 2.99--6.74\% with
$L_{\infty}$-norm. 

\begin{myblock}{Takeaways (Attack Strategies)}
Overall, across multiple setups, we observed that by applying the attack strategies we propose, adversaries can find
more attacks per time unit and obtain larger $\advantage{\impRel}{\allowedPert,\groundTruth,\classifier,\instanceSetGen,\impersonationGoalSet}(\adversary)$.
\end{myblock}

\section{Enhancing Group-based Robustness}
\label{sec:defenses}
We previously showed that our new metric can reveal new insights about
models' susceptibility to realistic threats (\secref{sec:metric}), as
well as that formalizing these threats makes it possible to design
faster or more successful attacks (\secref{sec:attacks}).  In this
section we introduce defenses that focus on defending against
group-based attacks and examine their performance relative to previous
defenses.
We summarize the desired properties of such defenses in \secref{sec:defenses:goal}.
We propose an approach to systematically build such defenses in \secref{sec:defenses:approach}.
We empirically verify that our defense achieves the desired properties in \secref{sec:defense:results},
with the setup described in \secref{sec:defense:setup}.

\subsection{Defense Objectives}
\label{sec:defenses:goal}
We desire the new defenses to have high benign accuracy on all inputs, and high $\robustness{\allowedPert,\groundTruth,\classifier,\instanceSetGen,\impersonationGoalSet}(\adversary)$,
preventing the attacker to achieve any $\impersonationGoal{\impersonationGoalIdx} \in \impersonationGoalSet$.
We also noticed that na\"{i}vely preventing the attacker from achieving their goal could
  also render the ML model useless in benign cases, when no attacker
  is present.  For example, by never classifying \emph{any} traffic sign as a stop sign,
  a classifier might achieve better group-based robustness because it might prevent unexpected vehicle stops caused by attacks. 
  Such a classifier might also maintain
  high average accuracy because there are many traffic signs other
  than stop signs that could still be classified correctly. However,
  such a classifier might not be considered useful in practice.

Hence, we desire the new defenses to perform better than
existing defenses on the following three metrics simultaneously:
1) group-based robustness $\robustness{\allowedPert,\groundTruth,\classifier,\instanceSetGen,\impersonationGoalSet}(\adversary)$,
 2) average accuracy, and 3) accuracy on unperturbed inputs associated
 with classes that might be impersonated.
When such classes are underrepresented, (2) might not naturally imply (3):
 (2) can still hold if there are
  many classes, of which only a few might be impersonated, and the
  classifier never emits them.

\subsection{Defense Approach}
\label{sec:defenses:approach}
The state-of-the-art defense against evasion attacks is adversarial training, 
which involves rapidly and adaptively generating
adversarial examples.\mahmood{these last two sentences don't read
  well, please revise}\weiran{Attempted} The more time-efficient group-based 
attacks that we have developed  (\secref{sec:attacks:loss}) make it possible to
generate adversarial examples fast enough for 
adversarial training specifically against group-based attacks. 

Existing adversarial training defenses train for better
untargeted robustness,
emphasizing that the model should always predict the \emph{correct
class}~\cite{iclr18:PGD, NeurIPS19:FreeTraining}. However, to
have high group-based robustness, a model does not have to be
always correct on perturbed (i.e., adversarial) inputs. Hence, we propose
focusing on training models to maintain accuracy on all inputs when no
attacker is present, while also allowing them to misclassify inputs
supplied by the attacker, as long as those misclassifications do not
further the attacker's objectives. For example, if an attacker, Eve,
attempts to impersonate a member of the bank staff, there is no harm
to the bank if the classifier is fooled into thinking that the
attacker is \emph{another} attacker, Mallory; the only harm is in
predicting that Eve is a member of the bank staff.

We believe this
approach will enable the achievement of higher group-based robustness while
maintaining high benign accuracy. 
We modified established adversarial training algorithms to build
stronger defenses against group-based attacks.
Besides generating adversarial examples with group-based attacks,
we also designed a new data-fetching mechanism and new loss function, detailed below.

\paragraph{Loss Function}
Many existing adversarial training defenses use the same
loss function for training as when no adversaries are
present (e.g., \cite{iclr18:PGD, NeurIPS19:FreeTraining}).
Intuitively, to have high group-based robustness, we only need to
train the model to avoid predicting any $\target \in \targetclassset$
on adversarial examples, and thus we have  
  \begin{equation}
\ell_{\mdtrain}=\advtrainweight*\sum_{\targetclass \in \targetclassset}ReLU(Z_{\targetclass}+\delta-\max_{i \notin \targetclassset}(Z_{i}))
\end{equation}
where $\targetclass$ is iterated over all classes in $\targetclassset$ and the
largest $Z_{i}$ among all classes \textit{i} not in $\targetclassset$ is used. 
$\advtrainweight$ is a non-negative weighting factor. 

Instinctively\mahmood{``intuitively'' is used too frequently}\weiran{Changed}, a larger $\advtrainweight$ implies higher group-based robustness and lower benign accuracy, and vice versa.
We realized that the choice of $\advtrainweight$ could depend on
the specific needs of implementation scenarios. For example, if we expect
the model to maintain very high benign accuracy while gaining some
group-based robustness, a large $\advtrainweight$ might be
preferred. In our implementation, we ran linear search
and chose a $\advtrainweight$ value for each dataset
such that our defense outperforms existing ones on all three metrics listed in \secref{sec:defenses:goal} on a validation set.

$\ell_{\mdtrain}$ is non-negative, and equals zero
if and only if an attack has been prevented. If
$\ell_{\mdtrain}>0$, there is at least one class $\targetclass
\in \targetclassset$ that has higher logits than all classes $i \notin
\targetclassset$; otherwise, if $\ell_{\mdtrain}=0$, there is at least
one class $i \notin
\targetclassset$ that has higher logits
than all classes $\targetclass
\in \targetclassset$.

\paragraph{Data Fetching}
Existing adversarial training defenses use adversarial examples generated from inputs associated with all classes. 
To have high group-based robustness, we only need to train the model
against evasion attacks that use input instances associated with some
$\sourceclass$ in $\sourceclassset$. 
As described in \secref{sec:defenses:goal}, we also expect the model to maintain high benign accuracy. 
Thus, we modify the data-fetching process so every fetched batch consists of two data partitions:
the first partition consists of inputs associated with all classes
and the second partition consists of inputs associated with some
$\sourceclass$ in $\sourceclassset$. In our implementation, the ratio
of sizes of partitions is close to the ratio of the inputs of these
two types in the training set. The first partition is never perturbed,
and we generate adversarial examples on the second partition using
$\ell_{\mdgroup}$ (\secref{sec:attacks:loss}).  
We train the model to correctly classify inputs in the first
partition, by minimizing the cross-entropy loss which is commonly used
for benign training. We train the model to prevent group-based attacks
on the second partition by minimizing $\ell_{\mdtrain}$. 

\subsection{Experiment Setup}
\label{sec:defense:setup}
To verify that our defense boosts
$\robustness{\allowedPert,\groundTruth,\classifier,\instanceSetGen,\impersonationGoalSet}(\adversary)$
at the same level of benign accuracy, we used the same threat model,
datasets and benchmarks as we did in \secref{sec:metric:setup}. In
addition, we used the baselines and measurement process below. 

\subsubsection{Baselines}
To illustrate its improvements to
$\robustness{\allowedPert,\groundTruth,\classifier,\instanceSetGen,\impersonationGoalSet}(\adversary)$
at the same level of benign accuracy, we examined the defense on the
\gtsrb{} dataset at $L_{\infty}=8/255$ with ResNet architecture, and
also on the \pubfig{} dataset. We reused the 100 adversarial training
instances as baselines on \gtsrb{}, and the VGG model trained by Wu et
al.'s method~\cite{iclr20:Wu2020Defending} for five epochs as a baseline on \pubfig{} (they were used in \secref{sec:metric:setup}). 

\subsubsection{Measurement Process}
As we did in \secref{sec:metric:setup}, \secref{sec:attacks:loss:process}, and \secref{sec:attacks:strategies:process},
we implemented \instanceSetGen 
  for each test $\impersonationGoalSet$ on different datasets. 
On the \gtsrb{} dataset, we used the same setup as we did for attack loss functions,
 perturbing speed limit and delimits
signs to (1) one of the
signs that would lead to an immediate stop or (2) no more than half of the actual limit. $\impersonationGoalSet =
\bigcup_{\sourceclass \in \sourceclassset} \bigcup_{\targetclass \in
  \targetclassset_{\sourceclass}} \{\{(\sourceclass, \targetclass)\}\}$.
We trained an instance with free adversarial training for 50 epochs,
and then trained it with the method we described in
\secref{sec:defenses:approach} for another 50 iterations.  
Other choices of numbers of iterations might also work.
We compared the robustness of this instance with the robustness of the
100 baselines, on adversarial examples generated by the best guess
method. 
  
Again we used the \pubfig{} dataset to
simulate the burglary scenario where burglars are trying to attack a bank.
Due to limited computation resources, we slightly modified the setup such that we still randomly selected two mutually exclusive
sets of classes $\sourceclassset$ and $\targetclassset$, but each of
$\sourceclassset$ and $\targetclassset$ consists of five
classes. 
We
randomly selected four different $\sourceclassset$ and 
$\targetclassset$ pairs. 
$\targetclassset$ is the set of staff members who have legal
access,
and $\sourceclassset$ is the set of burglars. 
As described in \secref{sec:intro},
the bank requires three \emph{distinct} staff members to
agree before a vault can be opened.  
The burglars need to impersonate three \emph{distinct} staff members and 
each $\impersonationGoal{\impersonationGoalIdx} \in \impersonationGoalSet$ is a surjective function.
We started with an instance that has been adversarially trained for
five epochs using Wu et al.'s method~\cite{iclr20:Wu2020Defending},
and we trained with our method (described in \secref{sec:defenses:approach}) for one epoch.
We also trained the VGG model
using Wu et al.'s method for one more epoch (six epochs in total) to fairly
compare with our defense. 
As described in \secref{sec:defenses}, our defenses are specific to $\impersonationGoalSet$, corresponding to choices of $\sourceclassset$ and $\targetclassset$.
We trained four instances according to the four choices.
To measure $\robustness{\allowedPert,\groundTruth,\classifier,\instanceSetGen,\impersonationGoalSet}(\adversary)$, we exhaustively searched through all possible choices of $ \instanceSet$:
we fixed
 $\vert \instanceSet \vert=5$, and \instanceSetGen sampled one $\instance$ from each $\sourceclass \in \sourceclassset$ uniformly. 
With respect to the $ \instanceSet$ and $\targetclassset$, we
exhaustively search through all possible choices of
$\impersonationGoal{\impersonationGoalIdx}$ and verify if there is
such an $\impersonationGoal{\impersonationGoalIdx}$ that the burglars
can achieve, as we described in \secref{sec:metric:definition}. It is
worth noticing that our defense aims to prevent attacks from happening
rather than make attacks slower (but only against the attack
strategies mentioned in \secref{sec:attacks:strategy}). 

As we explained in \secref{sec:defenses:goal}, we expect the model to
also maintain high accuracy on classes that might be impersonated. We
measured the benign accuracy on inputs associated with all possible
$\targetclass$ in $\targetclassset_{\sourceclass}$. On the \gtsrb{}
dataset, we measured the accuracy on any inputs $\instance$ that are
stop signs, no entry signs, no vehicle signs, or speed limits no
higher than 60 KPH. On the \pubfig{} dataset, we measured the accuracy
on inputs associated with any $\targetclass$ in $\targetclassset$ with
respect to the specific choices of $\targetclassset$. 

\subsection{Results}
\label{sec:defense:results}
\begin{figure}[t!]
\centerline{\includegraphics[width=0.95\columnwidth]{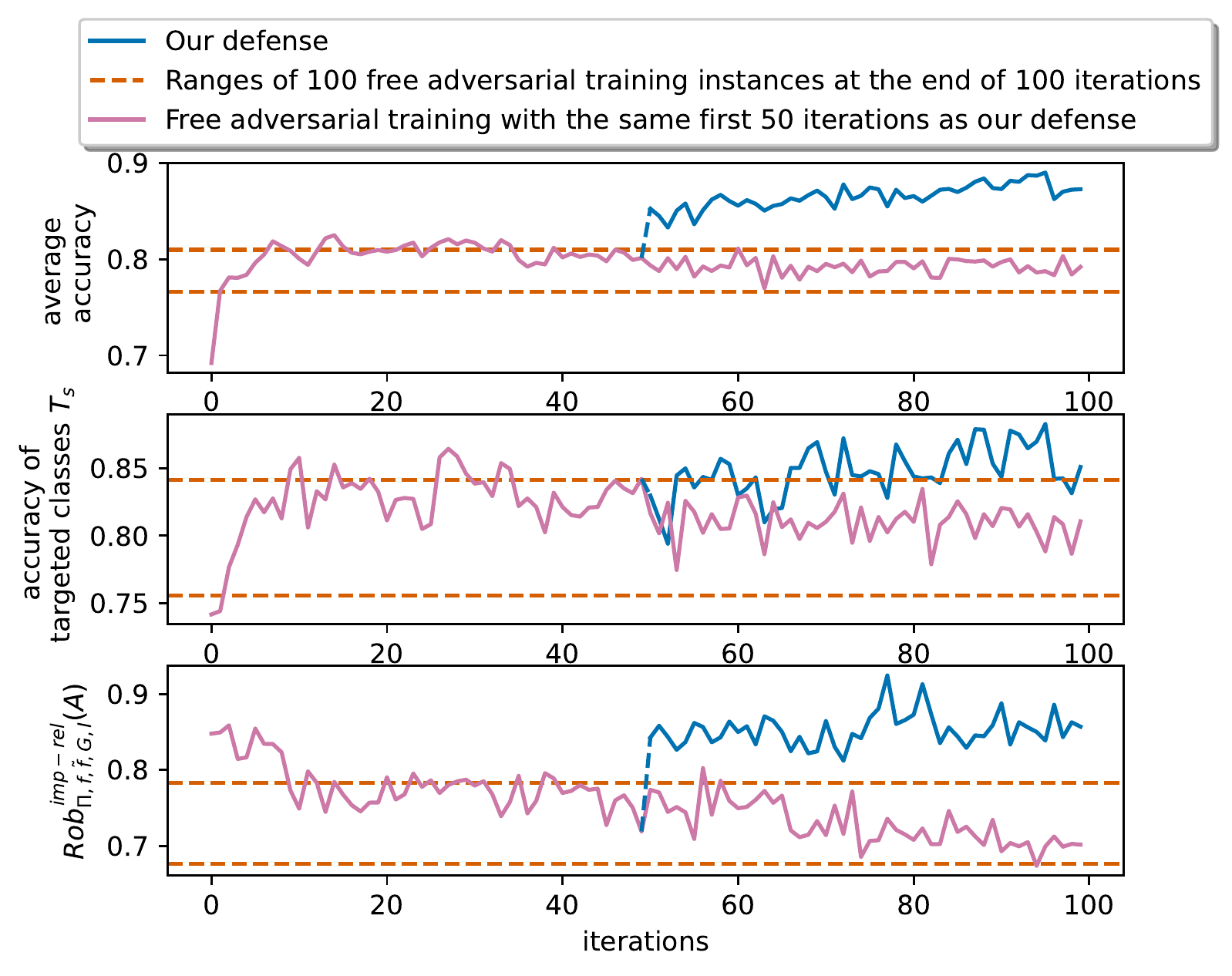}}
\caption{Average accuracy, accuracy on targeted classes
  $\targetclassset_{\sourceclass}$, and
  $\robustness{\allowedPert,\groundTruth,\classifier,\instanceSetGen,\impersonationGoalSet}(\adversary)$
  of our defense and free adversarial training. As soon as we switched
  to our defense, average accuracy and
  $\robustness{\allowedPert,\groundTruth,\classifier,\instanceSetGen,\impersonationGoalSet}(\adversary)$
  boosted while accuracy on $\targetclassset_{\sourceclass}$ remained
  about the same. As we trained for more iterations, the accuracy on
  $\targetclassset_{\sourceclass}$ and average accuracy kept
  increasing, while
  $\robustness{\allowedPert,\groundTruth,\classifier,\instanceSetGen,\impersonationGoalSet}(\adversary)$
  remained at the same level.} 
\label{fig:result:GTSRBdefense}
\end{figure}

\begin{figure}[t!]
\small
\caption{Peformance of our defense compared with its two baselines on the \pubfig{} dataset. 
Each instance of our defense corresponds to a specific choice of $\sourceclassset$ and $\targetclassset$.
Our defenses achieve similar average accuracy and  accuracy on
$\targetclassset$ to the baselines. Specifically, with $\#1$ choice of
$\targetclassset$, although our defense is up to $5\%$ less accurate
on inputs associated with $\targetclassset$, we are only two data
points worse because there are only $38$ input instances associated
with $\targetclassset$ in the test set. Meanwhile, the
$\robustness{\allowedPert,\groundTruth,\classifier,\instanceSetGen,\impersonationGoalSet}(\adversary)$
of our defense is much higher than the baselines. We measured
$\robustness{\allowedPert,\groundTruth,\classifier,\instanceSetGen,\impersonationGoalSet}(\adversary)$
with two choices of $\impersonationGoalSet$ as we did in
\secref{sec:attacks:strategies:results}. One choice of
$\impersonationGoalSet$ allows the reuse of attackers, a.k.a. the same
burglar \emph{may} impersonate more than one staff members, whereas
the other choice of $\impersonationGoalSet$ does not.} 
\begin{center}
\begin{tabular}{|@{\hskip 0.01in}c@{\hskip 0.01in}|@{\hskip 0.01in}c@{\hskip 0.01in}|@{\hskip 0.01in}r@{\hskip 0.01in}r@{\hskip 0.01in}@{\hskip 0.01in}|@{\hskip 0.01in}r@{\hskip 0.01in}r@{\hskip 0.01in}r@{\hskip 0.01in}r@{\hskip 0.01in}|}
\hline

&\textit{Choice}&\textit{Wu et}&\textit{Wu et}&\textit{Our}&\textit{Our}&\textit{Our}& $\textit{Our}$\\
\textit{Metric}&\textit{of}&\textit{al.'s}&\textit{al.'s}&\textit{Defense}&\textit{Defense}&\textit{Defense}& $\textit{Defense}$\\
&$\mathit{\targetclassset}$&&$\mathit{+1}$&$\mathit{\#1}$&$\mathit{\#2}$&$\mathit{\#3}$& $\mathit{\#4}$\\
&&&\textit{epoch}&&&& \\
\hline
Average&&&&&&& \\
Accuracy&-&0.98&0.98&0.98&0.99&0.99&0.99 \\
\hline
Accuracy&$\mathit{\#1}$&0.95&0.97&0.92&-&-&- \\
on&$\mathit{\#2}$&0.99&0.99&-&0.99&-&- \\
$\mathit{\targetclassset}$&$\mathit{\#3}$&1.00&0.98&-&-&1.00&- \\
&$\mathit{\#4}$&0.95&1.00&-&-&-&0.97 \\
\hline
Robustness&$\mathit{\#1}$&0.25&0.31&0.89&-&-&- \\
\emph{With}&$\mathit{\#2}$&0.27&0.38&-&0.65&-&- \\
Reuse of&$\mathit{\#3}$&0.84&0.55&-&-&0.97&- \\
Attackers&$\mathit{\#4}$&0.73&0.68&-&-&-&0.84 \\
\hline
Robustness&$\mathit{\#1}$&0.38&0.37&0.99&-&-&- \\
\emph{Without}&$\mathit{\#2}$&0.72&0.73&-&0.91&-&- \\
Reuse of&$\mathit{\#3}$&0.91&0.88&-&-&0.99&- \\
Attackers&$\mathit{\#4}$&0.95&0.98&-&-&-&1.00 \\
\hline
\end{tabular}
\end{center}
\label{tab:results:defense:glasses}
\end{figure}

\figref{fig:result:GTSRBdefense} demonstrates the results on the
\gtsrb{} dataset.  With one iteration of training, our defense
achieved higher average accuracy, higher
$\robustness{\allowedPert,\groundTruth,\classifier,\instanceSetGen,\impersonationGoalSet}(\adversary)$
and about the same accuracy on $\targetclassset_{\sourceclass}$
compared to the baselines. As the training process progressed, average accuracy and 
accuracy on $\targetclassset_{\sourceclass}$ increased,
while
$\robustness{\allowedPert,\groundTruth,\classifier,\instanceSetGen,\impersonationGoalSet}(\adversary)$
were always higher than that of the baselines.
On the
\pubfig{} dataset, as shown in \figref{tab:results:defense:glasses}, with one iteration of training, our defenses
achieve similar average accuracy and accuracy on $\targetclassset$ to
the baselines. Meanwhile, our defenses boost
$\robustness{\allowedPert,\groundTruth,\classifier,\instanceSetGen,\impersonationGoalSet}(\adversary)$
by up to $3.52$ times, relatively.  On both datasets, our defense has
no worse average accuracy and accuracy on $\targetclassset$ than the
baselines while obtaining higher
$\robustness{\allowedPert,\groundTruth,\classifier,\instanceSetGen,\impersonationGoalSet}(\adversary)$.
We acknowledge that our tuning of parameters might not be the best:
other tunings might achieve higher performance by some or all of the
three metrics mentioned above.  However, our experiments successfully demonstrated
that we can systematically generate defenses to meet the goals described in
\secref{sec:defenses:goal}.

\begin{myblock}{Takeaways (Defense)}
  By modifying existing adversarial training algorithms, we were able to generate defenses that outperform existing ones on all three metrics
  mentioned in~\secref{sec:defenses:goal}: 1) group-based robustness $\robustness{\allowedPert,\groundTruth,\classifier,\instanceSetGen,\impersonationGoalSet}(\adversary)$,
  2) average accuracy, and 3) accuracy on unperturbed inputs associated
 with classes that might be impersonated.
\end{myblock}

\section{Related Work}
\label{sec:relatedwork}

Several instances of prior work were discussed in \secref{sec:metric} and
\secref{sec:attacks}, which motivated the group-based metric and attacks. 
Additionally, \secref{sec:metric:setup} leverages prior
work as benchmarks and baselines. This section complements
the prior work already addressed by positioning group-based
 attacks within multiple domains, including
role-based access control, 
natural
perturbations, privacy, and fairness. We discuss each of
these in turn.


\subsubsection*{Role-based Access Controls}
Ferraiolo and Kuhn defined Role-based Access
Controls as a mechanism to group users by roles and grant access
accordingly~\cite{NCSC92:RBAC}. In a hospital setting, for example,
doctors have read and write access to prescriptions but pharmacists
only have read access. Doctors and pharmacists are two groups of
users, and doctors have more access.  Schaad et al.\ described a
real-world implementation of Role-based Access Controls in Dresdner
Bank, a  
major European bank where 50,659 employees are grouped into about 1,300
roles~\cite{SACMAT01:Bank}.  In the setup of our experiments, we also have two groups of users:
potential attackers and potential targets, e.g., students and instructors in the
class materials theft scenario, and burglars and bank staff in the burglary scenario. In both scenarios, users from groups with less
access attempt to impersonate members from the other groups to gain more
access. The new metric we propose, \gbtrfull, measures the true threat
of such disguise more accurately than existing metrics. The new
loss functions (\secref{sec:attacks:loss}) and strategies
(\secref{sec:attacks:strategy}) we propose help estimate this threat
more efficiently than than na\"{i}ve methods.

\subsubsection*{Natural Perturbations}
Machine-learning models have been found to perform
similarly~\cite{ICCCN:Humans} or better~\cite{Scientometrics20:Humans}
than humans on image-classification tasks.  However, when there is
noise in the images, these models tend to perform much worse than
humans~\cite{ICCCN:Humans}. For example, Mu et al.\ and Hendrycks et al.\ found that
natural, non-adversarial, perturbations, such as blurring, can significantly harm the
functionality of models~\cite{iclr19:ImageNetC,arxiv19:MNISTC}. Similarly, machine learning models' vulnerability to
group-based misclassifications demonstrated in this work could still exist even
without the presence of an adversary.

\subsubsection*{Evasion Attacks as a Defense for Privacy}
The prevalence of machine learning raises privacy concerns, spurring efforts to protect private information and resist surveillance.
Wenger et al.\ identified various cases where
facial recognition is deployed~\cite{oakland23:Sok} along with many anti-facial
recognition tools, including evasion
attacks (e.g.,~\cite{ccs16:eyeglasses,PETS22:face,PETS21:face,cvpr20:face,icpr21:face,cvpr19:face,ICCV20:face,ICIP19:face}).
Abdullah et al.\ summarized real-world use cases of
voice recognition systems~\cite{oakland21:Sok}, as well as corresponding evasion attacks against these
systems (e.g.,~\cite{ICML19:Speech,NDSS19:Speech,IJCAI19:Speech,Usenix20:Speech,Usenix20:Speech2,Usenix16:Speech,oakland21:Speech,Usenix18:Speech,NDSS19:Speech2}). As an introduction to a more general form of evasion based on groups (\secref{sec:metric}), our paper also provides 
a framework for evasion attacks that could similarly be used by individuals as countermeasures against unwanted surveillance or data collection (i.e., an individual from an oppressed or vulnerable minority group may want to prevent themselves from being automatically identified via facial recognition).

\subsubsection*{Fairness in Machine Learning}
In our experiments, the \gbtrfull of various models was significantly
affected by
different choices of $\targetclassset$ and $\sourceclassset$ even when we
used the same $\vert \targetclassset \vert$ and $\vert \sourceclassset
\vert$. We also observed that the success rate of perturbing images from
$\sourceclass \in \sourceclassset$ as $\target \in \targetclassset$
 differed significantly based on choices of
$\sourceclass$ and $\targetclass$, which led us to suggest
computing a prior probability of success for each $(\sourceclass,\target)$ to identify vulnerable
$(\instance,\target)$ pairs, with  $\instance \in \instanceSet$ (\secref{sec:attacks:strategy}). 
Researchers
have noticed a similar problem with accuracy and untargeted
robustness: the performance of models varies with different class
distribution~\cite{ICML21:fairness,FACCT:fairness}.
Previous work argues for 
fairness in machine learning, advocating that accuracy and untargeted
robustness should be independent of classes. However, whether a
counterpart of fair machine learning is feasible for targeted
robustness or \gbtrfull remains unknown.

\section{Conclusion}
\label{conclusion}
In this paper, we identified a limitation in the previous evaluation
process of defenses against evasion attacks: in some real-world attack
scenarios, the performance of these models cannot be accurately
measured by existing metrics. We formally defined a new metric, \gbtrfull, to measure
the true threat in these attack scenarios, and statistically verified
that \gbtrfull is negligibly or weakly correlated with every
existing metric. We also proposed approaches that, while maintaining a
close
$\advantage{\impRel}{\allowedPert,\groundTruth,\classifier,\instanceSetGen,\impersonationGoalSet}(\adversary)$,
boost the speed of attacks in these attack scenarios: two new loss
functions, \mdmaxloss and \mdgrouploss, and three new attack
strategies. 
We additionally innovated a defense that elevates \gbtrfull  $\robustness{\allowedPert,\groundTruth,\classifier,\instanceSetGen,\impersonationGoalSet}(\adversary)$ while maintaining high benign accuracy.
We validated the improvement with experiments across
datasets, defenses, distance metrics, and attack scenarios. Overall, we
explored a new attack space where some real-world attacks reside but
existing research works have not addressed.

\section*{Acknowledgments}
The work described in this paper was supported in part
by NSF grants 1801391, 2112562, and 2113345; 
by DARPA under contract HR00112020006;  
by the National Security Agency under award H9823018D0008; 
by Len Blavatnik and the Blavatnik Family Foundation;
by a Maof prize for outstanding young scientists;
and by the Neubauer Family Foundation.

\bibliographystyle{IEEEtran}
\bibliography{reference.bib}

%
%
\section{Artifact Appendix}

On behalf of the authors, we are happy to publish the source code of our paper as an artifact. However, our experiments took several months on the GPUs, while we are told by the artifact chairs that artifact reviewers are not assumed to have GPUs and artifacts to be reviewed are expected to take no longer than 24 hours. Thus we 1) decide to apply for the ``Functional" and ``Available" badges but not the ``Reproduced" badge and 2) provide a ``hello-world" style mini-experiment that takes less than two hours on laptop CPUs but still verifies our claims. 

\subsection{Description \& Requirements}

This section lists all the information necessary to recreate the experimental setup.

\subsubsection{How to access}
Our implementation is stored in this public GitHub repository: \url{https://github.com/linweiran/GBR}. The experiments we specifically designed for artifact evaluation can be found in the GTSRB repository, a.k.a. \url{https://github.com/linweiran/GBR/tree/main/GTSRB}.
We have created a Zenodo version at \url{https://zenodo.org/records/10104298}, with the DOI: 10.5281/zenodo.10104297.

\subsubsection{Hardware dependencies}
The experiments we propose in the artifact would take less than two hours on laptop CPUs, so there is no other hardware dependency.
\subsubsection{Software dependencies} 

\label{Software-dependencies}
We recommend using python3 of version 3.10.9 to run our scripts. Additionally, we require a list of python packages to run our code. The specified list of python packages can be found at \url{https://github.com/linweiran/GBR/blob/main/GTSRB/requirements-cpu.txt}. To install all packages, you may easily run ``pip3 install -r requirements-cpu.txt" within the GTSRB folder. 

\subsubsection{Benchmarks} 
\label{artifact:benchmarks}
The dataset we used in the artifact is the GTSRB dataset. Please download the dataset from the downloads section of \url{https://benchmark.ini.rub.de/gtsrb_dataset.html#Downloads} where a link can be found. Specifically, please download files named ``GTSRB\_Final\_Training\_Images.zip", ``GTSRB\_Final\_Test\_Images.zip", and ``GTSRB\_Final\_Test\_GT.zip". After extracting these zip files, please move the directories named ``Final\_Test" and ``Final\_Training", along with the file ``GT-final\_test.csv" to the same directory. The path of this directory will be used as the only parameter (for both scripts mentioned below (\secref{Artifact:Installation} and \secref{Experiment-Workflow})

\subsection{Artifact Installation \& Configuration}
\label{Artifact:Installation}

We preprocessed GTSRB images as the first step. You may switch to the GTSRB directory in our repo and run ``python3 preprocess.py --data\_path DATA\_PATH" where ``DATA\_PATH" is the path to the directory where the extracted files are stored (described in \secref{artifact:benchmarks}). An example could be  ``python3 preprocess.py --data\_path \textbackslash data\textbackslash GTSRB". The results will be printed out to the console.

\subsection{Experiment Workflow}
\label{Experiment-Workflow}

We wrapped up all the mini-experiments as a single script named ``hello\_world.py" under the GTSRB directory. To reproduce experiments, run ``python3 hello\_world.py --data\_path DATA\_PATH" where ``DATA\_PATH" is the path to the directory where the extracted files are stored (described in \secref{artifact:benchmarks} and \secref{Artifact:Installation}). An example could be  ``python3 hello\_world.py --data\_path \textbackslash data\textbackslash GTSRB".

\subsection{Major Claims}
\label{Major-Claims}
The major claims we made in the paper include:

\begin{itemize}
    \item (C1): \Gbtrfull $\robustness{\allowedPert,\groundTruth,\classifier,\instanceSetGen,\impersonationGoalSet}(\adversary)$ measures the robustness of models
  using different choices of $\impersonationGoalSet$ in accordance with
  the attack scenarios.
  Conventional metrics cannot measure the true threat in these
  sophisticated scenarios as accurately as \gbtrfull does. 
  Thus, we conclude that
  \gbtrfull offers another meaningful
    assessment of model susceptibility to attacks in the real world
    compared to conventional metrics (\S 2).
    \item (C2): Attacks with \mdmaxloss or \mdgrouploss
achieve comparable or slightly lower 
$\advantage{\impRel}{\allowedPert,\groundTruth,\classifier,\instanceSetGen,\impersonationGoalSet}(\adversary)$
than the best-guess attacks, consume markedly less time and are
markedly more efficient, finding more attacks per time unit. Attacks with \mdmaxloss or \mdgrouploss
 consume the same amount of time as the average-guess attacks and have much higher 
 $\advantage{\impRel}{\allowedPert,\groundTruth,\classifier,\instanceSetGen,\impersonationGoalSet}(\adversary)$. 
The MDMUL loss
and \mdmaxloss boost the efficiency of attacks in the attack
scenarios we tried (\S 3.A).
 \item (C3): By applying the attack strategies we propose, adversaries can find
more attacks per time unit and obtain larger $\advantage{\impRel}{\allowedPert,\groundTruth,\classifier,\instanceSetGen,\impersonationGoalSet}(\adversary)$ (\S 3.B).
 \item (C4): By modifying existing adversarial training algorithms, we were able to generate defenses that outperform existing ones on all three metrics: 1) group-based robustness $\robustness{\allowedPert,\groundTruth,\classifier,\instanceSetGen,\impersonationGoalSet}(\adversary)$,
  2) average accuracy, and 3) accuracy on unperturbed inputs associated
 with classes that might be impersonated 
(\S 4).
\end{itemize}

\subsection{Evaluation}
\label{Evaluation}

This section documents the details of our experiments. 
As we documented at the beginning of this appendix, 
we understand that artifact reviewers have much more limited computation resources than we have. Thus we are not applying for the ``Reproduced" badge and designed this scale-down set of mini-experiments. Specifically, we performed the following modifications to reduce experiments:
\begin{itemize}
 \item We ran mini-experiments only on a specific combination of settings. The full-scale experiments used multiple datasets, distance limits, model architectures, and random seeds. The min-experiments used only the GTSRB dataset, $L_{\infty}$ distance at $8/255$, SqueezeNet architecture, and $0$ as the random seed. We used such a combination as it is one of the least time-costly combinations.
 \item Mini-experiments ran all attacks by only one iteration. The full-scale experiments ran attacks with their default settings, ranging from $100$ to $300$ iterations.
  \item Mini-experiments trained models by as few iterations as possible (such that the models still converge). Full-scale experiments trained models by $100$ to $300$ iterations. while the mini-experiments only trained up to eight iterations. 
  \end{itemize}
 We acknowledge that compared to full-scale experiments, the mini-experiments used weaker attacks and worse-performing models (e.g. less robust defenses). However, the major conclusions we made (\secref{Major-Claims}) still hold on the mini-experiments. It is worth noticing that the mini-experiments and full-scale experiments call exactly the same functions: only the parameters (specified above) sent to these functions differ. 
The overall time of our experiments shall take less than two hours on laptop CPUs. 

\subsubsection{Experiment (E1)}
\textit{[Setup]}
As described in the main paper, attackers are perturbing speed limit and delimit signs  
 into signs that:
\begin{itemize} 
\item require an immediate stop, including stop signs, no-entry signs, and no-vehicle
signs; or
\item display a limit much lower than the actual limit, such as no more than half of the actual limit.
\end{itemize}

\textit{[How to]} 
We will document the steps required to prepare and configure the environment for this experiment, the steps to run this experiment, and the steps required to collect and interpret the results for this experiment in the following three blocks correspondingly.

\textit{[Preparation]}
First, install Python packages as we described in \secref{Software-dependencies}. We highly recommend the use of a virtual environment. Then download the dataset as we described in \secref{artifact:benchmarks}. Ultimately, preprocess the downloaded data as we described in \secref{Artifact:Installation}

\textit{[Execution]}
As we described in \secref{Experiment-Workflow}, we wrapped up all the mini-experiments as a single script named ``hello\_world.py". You may run ``python3 hello\_world.py --data\_path DATA\_PATH".

\textit{[Results]}
Four numbers will be reported: benign accuracy, untargeted robustness, targeted robustness, and group-based robustness. Group-based robustness is different from all the other three. This supports C1.

\subsubsection{Experiment (E2)}
\textit{[Setup]}
Same as the setup of E1.

\textit{[How to]} 
Same as the [How to] of E1.

\textit{[Preparation]}
Same as the [Preparation] of E1.

\textit{[Execution]}
Same as the[Execution] of E1.

\textit{[Results]}
Five numbers will be reported, which correspond to the success rates of five attacks: attacks with the MDMAX loss, attacks with the MDMUL loss, the best guess attacks, the worst guess attacks, and the average guess attacks. The success rate of attacks with the MDMAX loss or attacks with the MDMUL loss is higher than that of the worst guess attacks or the average guess attacks. This supports C2.

\subsubsection{Experiment (E3)}
\textit{[Setup]}
We did not evaluate strategies on the GTSRB dataset in the main paper.
Here the setup is that attackers are perturbing speed limits no less than 70 (five classes) as speed limits no higher than 60 (four classes).
Attackers sample one image from each of the five higher-speed classes, in total five images as a set.
For each set of five images, attackers can only claim success if they can manipulate these images as all of the lower-speed four classes. They may manipulate the same image as different signs.
We use the worst-performing strategy that we proposed (Estimate by Computing a Prior from a Validation Set).

\textit{[How to]} 
Same as the [How to] of E1.

\textit{[Preparation]}
Same as the [Preparation] of E1.

\textit{[Execution]}
Same as the[Execution] of E1.

\textit{[Results]}
Two numbers will be reported, which are the number of attempts needed by attackers to find the same number of successful attacks, with or without using our strategies. Attackers using our strategies need fewer attempts. This supports C3.

\subsubsection{Experiment (E4)}
\textit{[Setup]}
Same as the setup of E1 and E2.

\textit{[How to]} 
Same as the [How to] of E1.

\textit{[Preparation]}
Same as the [Preparation] of E1.

\textit{[Execution]}
Same as the[Execution] of E1.

\textit{[Results]}
Three sets of numbers are reported. Each set consists of three numbers: benign accuracy, benign accuracy on the targeted set, and group-based robustness. The three sets correspond to an existing defense, adversarial training (7 training iterations), adversarial training with one more iteration of our defense training (8 iterations in total), and adversarial training with one more iteration (8 iterations of adversarial training). The model with our defense outperforms the other two models in all three metrics. This supports C4.


\end{document}